%% file: arxiv.tex
\documentclass{article}

\usepackage[preprint]{neurips_2026}

\usepackage[utf8]{inputenc}
\usepackage[T1]{fontenc}

\usepackage{xcolor}
\definecolor{LinkBlue}{RGB}{0,70,140}
\definecolor{CiteGreen}{RGB}{0,120,80}
\definecolor{URLPurple}{RGB}{100,45,145}
\usepackage[
    colorlinks=true,
    linkcolor=LinkBlue,
    citecolor=CiteGreen,
    urlcolor=URLPurple
]{hyperref}

\usepackage{url}
\usepackage{booktabs}
\usepackage{multirow}
\usepackage{subcaption}
\usepackage{graphicx}
\usepackage{array}
\usepackage{tabularx}
\usepackage{longtable}
\usepackage{nicefrac}
\usepackage{microtype}

\usepackage{amsmath}
\usepackage{amsfonts}
\usepackage{amssymb}
\usepackage{amsthm}

\usepackage{algorithm}
\usepackage{algorithmic}

\usepackage{titletoc}
\newcommand\DoToC{%
  \startcontents
  \printcontents{}{1}{\hrulefill\vskip0pt}
  \vskip0pt \noindent\hrulefill
}

\newcolumntype{C}[1]{>{\centering\arraybackslash}m{#1}}
\newcolumntype{L}[1]{>{\raggedright\arraybackslash}p{#1}}
\newcolumntype{Y}{>{\raggedright\arraybackslash}X}

\definecolor{TaxBlue}{RGB}{33,82,145}
\definecolor{TaxGray}{RGB}{105,105,105}

\newcommand{\taxcat}[1]{\textcolor{TaxBlue}{\textbf{#1}}}
\newcommand{\formtag}[1]{\textcolor{TaxGray}{\footnotesize\textsc{#1}}}
\newcommand{\reftight}[1]{{\raggedright\footnotesize #1\par}}
\newcommand{\softmidrule}{\addlinespace[2pt]\midrule\addlinespace[2pt]}

\newcommand{\firstnotecat}[1]{%
  \multicolumn{3}{@{}l}{\hspace{0.6em}\textcolor{TaxBlue}{\textbf{#1}}}\\[-1pt]
  \midrule
}

\newcommand{\notecat}[1]{%
  \midrule
  \multicolumn{3}{@{}l}{\textcolor{TaxBlue}{\textbf{\textsc{#1}}}}\\[-1pt]
  \midrule
}

\title{A Dialogue between Causal and Traditional Representation Learning:
Toward Mutual Benefits in a Unified Formulation}

\author{%
\begin{tabular}{c}
Yan Li\textsuperscript{1} \quad
Yuewen Sun\textsuperscript{1,2}
 \quad
Shaoan Xie\textsuperscript{2} \quad
Gongxu Luo\textsuperscript{1} \quad
Yunlong Deng\textsuperscript{1} 
\\[2pt]
Kun Zhang\textsuperscript{1,2} \quad
Guangyi Chen\textsuperscript{1,2}
\\[8pt]
{\normalfont\mdseries\upshape
\textsuperscript{1}Mohamed bin Zayed University of Artificial Intelligence
}
\\[-1pt]
{\normalfont\mdseries\upshape
\textsuperscript{2}Carnegie Mellon University
}
\end{tabular}
}

\begin{document}

\maketitle

\begin{abstract}
Causal representation learning (CRL) and traditional representation learning have largely developed along different trajectories. Traditional representation learning has been driven mainly by applications and empirical objectives, whereas CRL has focused more on theoretical questions, particularly identifiability. This difference in emphasis has created a gap between the two fields in terminology, problem formulation, and evaluation, limiting communication and sometimes leading to disconnected or redundant efforts. In this paper, we argue that these two fields should be brought into dialogue rather than treated as separate paradigms. To this end, we introduce a unified formulation in which the representation learning is characterized by two components: a task component, which specifies what information the learned representation is required to preserve, and a constraint component, which specifies what structure is imposed on the latent space. 
Under this formulation, the benefits run in both directions. CRL provides theoretical tools for understanding when structured latent constraints are useful or necessary, while traditional representation learning offers practical insights on task design and objective choice that can improve the development of CRL methods. To illustrate this interaction, we experimentally study how different task components affect the behavior of CRL methods under different structured constraints. Results on CausalVerse show that the effectiveness of causal constraints depends strongly on the tasks with which they are paired.
\end{abstract}

\input{sections/1_introduction}

\input{sections/2_formulation_rl}
\input{sections/3_tradation_rl}

\input{sections/4_crl}

\input{sections/5_mutual_benifits}
\input{sections/6_experiments}

\input{sections/7_conclusion}

\bibliographystyle{unsrtnat}
\bibliography{ref}

\clearpage
\appendix

\begin{center}
\textit{\large Appendix for}\\[1em]
{\large \bfseries A Dialogue between Causal and Traditional Representation Learning: Toward Mutual Benefits in a Unified Formulation}
\end{center}

\newcommand{\beginsupplement}{%
\setcounter{table}{0}
\renewcommand{\thetable}{A\arabic{table}}%
\renewcommand{\theHtable}{A\arabic{table}}%
\setcounter{figure}{0}
\renewcommand{\thefigure}{A\arabic{figure}}%
\renewcommand{\theHfigure}{A\arabic{figure}}%
\setcounter{algorithm}{0}
\renewcommand{\thealgorithm}{A\arabic{algorithm}}%
\renewcommand{\theHalgorithm}{A\arabic{algorithm}}%
\setcounter{section}{0}
\renewcommand{\thesection}{A\arabic{section}}%
\renewcommand{\theHsection}{A\arabic{section}}%
\setcounter{theorem}{0}
\renewcommand{\thetheorem}{\arabic{theorem}}%
\renewcommand{\theHtheorem}{\arabic{theorem}}%
}

\beginsupplement

{\large Table of Contents:}

\DoToC
\clearpage

\input{appendix/1-notation}
\input{appendix/2-additional-details}
\input{appendix/3-experiment_details}


\end{document}

%% file: sections/1_introduction.tex
\section{Introduction}

Representation learning aims to extract latent representations that preserve information useful for downstream objectives such as prediction, classification, and detection. Over the past decade, this paradigm has been instantiated through a wide range of methods, including contrastive learning~\citep{oord2018cpc,he2020moco}, masked autoencoding~\citep{he2022mae}, and transformation prediction~\citep{gidaris2018rotation,noroozi2016jigsaw}. Recently, causal representation learning (CRL) has emerged as a related yet distinct line of research~\citep{scholkopf2021toward,yaotdrl,ahuja2023interventional}. CRL aims to learn latent representations that capture the underlying causal factors and their relations in the data. Existing work has made substantial progress on its theoretical foundations, especially on identifiability, and has therefore often focused on the assumptions under which causal representations can be recovered.

Despite this progress, CRL and traditional representation learning have developed along different trajectories. Traditional representation learning is largely driven by applications and empirical objectives, while CRL is more theory-driven. This divergence affects not only the problems the two communities emphasize, but also the language they use: traditional representation learning is often discussed in terms of objectives, augmentations, and empirical utility, whereas CRL is framed in terms of causal variables, structural equations, interventions, and identification. As a result, the two fields often approach related questions with different standards of success, limiting communication and sometimes leading to disconnected or redundant efforts.

In this paper, we study the relationship between the two fields from a learning perspective. We introduce a unified formulation with two components: a \emph{task component}, which specifies what information the learned representation is required to preserve, and a \emph{constraint component}, which specifies what structure is imposed on the latent space. Under this view, traditional representation learning mainly focuses on task components, possibly with generic regularization, whereas CRL is distinguished by structured constraints, such as conditional independence constraints, to align the learned representation with underlying causal mechanisms. This formulation provides a common language for both fields and makes their relationship more explicit.

Under this formulation, \textbf{we argue that the benefits run in both directions}. CRL provides theoretical tools for understanding when structured latent constraints are useful or necessary, especially for problems requiring reasoning about underlying generative structure, such as asymmetric cross-modal alignment~\citep{smartclip}, counterfactual reasoning~\citep{llcp}, and controllable generation~\citep{concept-aligner}. Conversely, traditional representation learning offers practical insights into task design and objective choice, which can improve CRL in real applications. In particular, our formulation highlights an underexplored aspect of current CRL research: existing work focuses mainly on structured constraints, while paying much less attention to the task objectives with which they should be combined. This suggests that \textbf{the practical behavior of CRL methods is shaped not by constraints alone, but by the combination of task and constraint components}.

To ground this perspective empirically, we conduct an experimental study on the CausalVerse benchmark. We compare multiple task formulations, including reconstruction-based, contrastive, and masked prediction objectives, and analyze how they affect learned representations under different structured constraints. Our results show that the effectiveness of causal constraints depends strongly on the task they are paired with, supporting the importance of task--constraint interaction.

Our contributions are threefold. First, we propose a unified formulation for describing both traditional representation learning and CRL in terms of a task component and a constraint component. Second, we use this formulation to clarify their relationship from a learning perspective, emphasizing how the two fields can mutually benefit each other. Third, we empirically study the role of the task component in CRL on CausalVerse and show that task--constraint combinations play a decisive role in practice.

%% file: sections/2_formulation_rl.tex
\section{A unified formulation of representation learning}
\label{sec:unified_framework}

We formulate representation learning as the joint specification of two components: a \emph{task component}, which determines what information the representation is required to preserve, and a \emph{constraint component}, which determines what structure is imposed on the latent space.

Let $X\in\mathcal{X}$ denote an observed sample drawn from a data distribution $p_{\mathrm{data}}(X)$. A learner does not necessarily observe $X$ directly. Instead, it receives a view
\begin{equation}
    \tilde{X}\sim {v}(X),
\end{equation}
where ${v}$ denotes a view-generation process, such as the identity map, stochastic augmentation, masking, corruption, temporal truncation, or the construction of paired views. An encoder $f_\theta$ maps the observed view to a latent representation
\begin{equation}
    Z=f_\theta(\tilde{X})\in\mathcal{Z}.
\end{equation}
A predictor or decoder $g_\phi$ then uses $Z$ to predict a target signal $T(X)$ induced by the learning objective. The general learning problem can be written as
\begin{equation}
\label{eq:unified_objective}
    \min_{\theta,\phi}\;
    \mathbb{E}_{X,\,\tilde{X}\sim {v}(X)}
    \left[
        \mathcal{L}_{\mathrm{task}}
        \big(g_\phi(Z),\,T(X)\big)
        +\lambda\,\Omega(Z,X)
    \right],
    \qquad
    Z=f_\theta(\tilde{X}).
\end{equation}
Here $\mathcal{L}_{\mathrm{task}}$ is the task loss, and $\Omega$ is an additional constraint on the learned latent space. The task component is specified by the view-generation process $\mathcal{V}$, the target signal $T(X)$, and the associated loss $\mathcal{L}_{\mathrm{task}}$; the constraint component is specified by the constraint function $\Omega$ and its weight $\lambda$.



%% file: sections/3_tradation_rl.tex
\section{Traditional representation learning under the unified formulation}
\label{sec:traditional_rl}

Traditional representation learning can be understood as a broad family of methods obtained from Eq.~\eqref{eq:unified_objective} by choosing different view-generation processes ${v}$, target signals $T(X)$, and generic constraints $\Omega$. We organize this section into two parts. Section~\ref{subsec:task_driven_rl} discusses task-driven representation learning, where $\Omega$ is absent or weak. Section~\ref{subsec:generic_constraints} discusses methods that supplement task-driven learning with generic latent constraints. 

\begin{table}[tbp]
\centering
\caption{\textbf{Task-driven representation learning under the unified formulation.} Each method is specified by an observed view $\tilde{X}$, a target signal $T(X)$, and a task loss $\mathcal{L}_{\mathrm{task}}$. Here $M^{\urcorner}$ denotes the visible complement of the masked set $M$, $\mathcal N^-$ denotes the set of negative candidates for sample $X$, $s(\cdot,\cdot)$ is a similarity function, such as the cosine similarity, and $q(X)$ denotes a cluster or prototype assignment.}
\vspace{6pt}
\small
\setlength{\tabcolsep}{4.5pt}
\renewcommand{\arraystretch}{1.20}
\begin{tabularx}{\textwidth}{@{}L{2.2cm}Y L{2.70cm}@{}}
\toprule
\textbf{Category} & \textbf{Formulation} & \textbf{Reference} \\
\midrule

\taxcat{Target prediction}
&
\formtag{View / target / representation}\quad
$\tilde{X}=X,\; T(X)=Y,\; Z=f_\theta(X)$

\vspace{2pt}
\formtag{Task loss}\quad
$
\mathcal{L}_{\mathrm{task}}
=
-\log p_\phi(Y\mid Z)
$
&
\reftight{\citep{lecun1998gradient,krizhevsky2012imagenet,he2016resnet,yosinski2014transfer}}
\\
\softmidrule

\taxcat{Reconstruction}
&
\formtag{View / target / representation}\quad
$\tilde{X}=X,\; T(X)=X,\; Z=f_\theta(X)$

\vspace{2pt}
\formtag{Task loss}\quad
$
\mathcal{L}_{\mathrm{task}}
=
\|X-g_\phi(Z)\|_2^2
$

\vspace{2pt}
\formtag{Variant}\quad
Denoising reconstruction uses $\tilde{X}=v^{cor}(X)$ while keeping $T(X)=X$.
&
\reftight{\citep{hinton2006reducing,vincent2008extracting,vincent2010stacked,masci2011stacked}}
\\
\softmidrule

\taxcat{Partial information prediction}
&
\formtag{View / target / representation}\quad
$\tilde{X}=X_{M^{\urcorner}},\; T(X)=X_M,\; Z=f_\theta(X_{M^{\urcorner}})$

\vspace{2pt}
\formtag{Task loss}\quad
$
\mathcal{L}_{\mathrm{task}}
=
\|X_M-g_\phi(Z)\|_2^2
$
&
\reftight{\citep{devlin2019bert,he2022mae,pathak2016context}}
\\
\softmidrule

\taxcat{Transformation correction}
&
\formtag{View / target / representation}\quad
$\tilde{X}=v(X),\; T(X)=v,\; Z=f_\theta(v(X))$

\vspace{2pt}
\formtag{Task loss}\quad
$
\mathcal{L}_{\mathrm{task}}
=
-\log p_\phi(v\mid Z)
$

\vspace{2pt}
\formtag{Examples}\quad
$v$ is the transformation, such as rotation $v^{rot}$ and jigsaw permutation $v^{jig}$.
&
\reftight{\citep{gidaris2018rotation,noroozi2016jigsaw,doersch2015context,pathak2016context,zhang2016colorization}}
\\
\softmidrule

\taxcat{Contrastive learning}
&
\formtag{View / target}\quad
$\tilde{X}=v_1(X),\; T(X)=v_2(X)$

\vspace{2pt}
\formtag{Projection}\quad
$
h=g_\phi(f_\theta(v_1(X))),\;
h^+=g_\phi(f_\theta(v_2(X)))
$

\vspace{2pt}
\formtag{Task loss}\quad
$
\mathcal{L}_{\mathrm{task}}
=
-\log
\frac{\exp(s(h,h^+)/\tau)}
{\exp(s(h,h^+)/\tau)+
\sum_{H\in\mathcal{N}^-}\exp(s(h,H)/\tau)}
$
&
\reftight{\citep{oord2018cpc,he2020moco,chen2020simclr,grill2020bootstrap,chen2021exploring,zbontar2021barlow}}
\\
\softmidrule

\taxcat{Cluster / prototype learning}
&
\formtag{View / target / representation}\quad
$\tilde{X}=X,\; T(X)=q(X),\; Z=f_\theta(X)$

\vspace{2pt}
\formtag{Task loss}\quad
$
\mathcal{L}_{\mathrm{task}}
=
-\log p_\phi(q(X)\mid Z)
$
&
\reftight{\citep{xie2016unsupervised,caron2018deepcluster,caron2020unsupervised,ji2019invariant}}
\\

\bottomrule
\end{tabularx}
\label{tab:task_driven_rl}
\end{table}

\subsection{Task-driven representation learning}
\label{subsec:task_driven_rl}

When the constraint term is absent or only plays a weak regularizing role, Eq.~\eqref{eq:unified_objective} reduces to
\begin{equation}
\label{eq:task_driven_objective}
    \min_{\theta,\phi}\;
    \mathbb{E}_{X,\,\tilde{X}\sim{v}(X)}
    \left[
        \mathcal{L}_{\mathrm{task}}
        \big(g_\phi(f_\theta(\tilde{X})),\,T(X)\big)
    \right].
\end{equation}
Under this view, representation learning methods differ mainly in what the learner observes and what target signal the representation is trained to preserve.

This task-driven view yields a taxonomy based on the predictive signal used to shape the representation.
\textbf{Target prediction} uses an external label $Y$ as the target signal. 
\textbf{Reconstruction} uses the raw input itself as the target. Its variant, denoising reconstruction, changes the observed view to a corrupted input $v^{cor}(X)$. 
\textbf{Partial information prediction} observes the partial visible portion $X_{M^{\urcorner}}$ and predicts the missing portion $X_M$. 
\textbf{Transformation correction} constructs the target from the transformation $v$ applied to the input, such as a rotation or patch permutation. Usually, the target is classifying the transformation or reconstruction from transformed inputs. 
\textbf{Contrastive learning} uses paired views $v_1(X)$ and $v_2(X)$, and trains the representation to identify the positive view among candidate views.
\textbf{Cluster or prototype learning} defines the target as an assignment $q(X)$ induced from the current representation space. 
These objectives instantiate different choices of $\tilde{X}$ and $T(X)$ within the task-driven objective in Eq.~\eqref{eq:task_driven_objective}. 
Table~\ref{tab:task_driven_rl} gives the corresponding formulations. Please note that we do not attempt to cover all variants within each category, and instead include only a selection of representative methods here.
More details are provided in Appendix~\ref{app:task_driven_rl}.

\begin{table}[tbp]
\centering
\caption{\textbf{Representation learning with generic latent constraints.} These methods supplement a task component with a non-causal constraint $\Omega_{\mathrm{gen}}$ that shapes the latent space through compression, prior alignment, sparsity, energy-based preferences, or discreteness. $q_\theta$ denotes a stochastic encoder, $p$ denotes a reference distribution, $D_{\mathrm{KL}}(\cdot\|\cdot)$ denotes KL divergence, $E_\psi$ denotes an energy function, and $\{e_k\}_{k=1}^{K}$ denotes a finite codebook. }
\vspace{6pt}
\label{tab:generic_constraints}
\small
\setlength{\tabcolsep}{4.5pt}
\renewcommand{\arraystretch}{1.20}
\begin{tabularx}{\textwidth}{@{}L{2cm}Y L{3cm}@{}}
\toprule
\textbf{Category} & \textbf{Formulation} & \textbf{Reference} \\
\midrule

\taxcat{Compression constraints}
&
\formtag{Setting}\quad
$Z\sim q_\theta(Z\mid X),\; T(X)=Y$

\vspace{2pt}
\formtag{Objective}\quad
$
\mathcal{L}
=
-\log p_\phi(Y\mid Z)
+\beta I(X;Z)
$

\vspace{2pt}
\formtag{Variant}\quad
$I(X;Z)$ is replaced by
$D_{\mathrm{KL}}(q_\theta(Z\mid X)\|p(Z))$.
&
\reftight{\citep{tishby2000ib,alemi2017vib,achille2018information}}
\\
\softmidrule

\taxcat{Distribution alignment constraints}
&
\formtag{Setting}\quad
$Z\sim q_\theta(Z\mid X),\; T(X)=X$

\vspace{2pt}
\formtag{Objective}\quad
$
\mathcal{L}
=
-\log p_\phi(X\mid Z)
+
\beta D_{\mathrm{KL}}
(q_\theta(Z\mid X)\|p(Z))
$

\vspace{2pt}
\formtag{Variant}\quad
KL term can be replaced with adversarial learning.
&
\reftight{\citep{kingma2014vae,rezende2014stochastic,makhzani2016aae}}
\\
\softmidrule

\taxcat{Sparsity constraints}
&
\formtag{Setting}\quad
$Z=f_\theta(X),\; T(X)=X$

\vspace{2pt}
\formtag{Objective}\quad
$
\mathcal{L}
=
\|X-g_\phi(Z)\|_2^2
+\lambda\|Z\|_1
$

\vspace{2pt}
\formtag{Variant}\quad
L1 term can be 
replaced by the KL divergence with target sparsity level $\rho$:
$
D_{\mathrm{KL}}(\rho\|\hat{\rho})
$
&
\reftight{\citep{olshausen1996emergence,olshausen1997sparse,ng2011sparseae,cunningham2023sparse,templeton2024scaling}}
\\
\softmidrule

\taxcat{Energy-based constraints}
&
\formtag{Prior}\quad
$ Z=f_\theta(X),\qquad
p(Z)\propto \exp(-E_\psi(Z))
$

\vspace{2pt}
\formtag{Constraint}\quad
$
\Omega_{\mathrm{gen}}(Z,X)=E_\psi(Z)
$

\vspace{2pt}
\formtag{Objective}\quad
$
\mathcal{L}
=
\|X-g_\phi(Z)\|_2^2
+\lambda E_\psi(Z)
$
&
\reftight{\citep{lecun2006ebm,hinton2006dbn,hinton2012rbm,bengio2007greedy}}
\\
\softmidrule

\taxcat{Discrete / quantized constraints}
&
\formtag{Quantization}\quad
$
z_e=f_\theta(X),\quad
Z=e_{k^\ast},\quad
k^\ast=\arg\min_k\|z_e-e_k\|_2^2
$

\vspace{2pt}
\formtag{Objective}\quad
$
\mathcal{L}
=
\|X-g_\phi(Z)\|_2^2
+\|\mathrm{sg}[z_e]-e_{k^\ast}\|_2^2
+\beta\|z_e-\mathrm{sg}[e_{k^\ast}]\|_2^2
$
&
\reftight{\citep{oord2017vqvae,razavi2019vqvae2,esser2021vqgan, yu2023language, esser2021taming}}
\\

\bottomrule
\end{tabularx}
\vspace{-0.3cm}
\end{table}

\subsection{Representation learning with generic constraints}
\label{subsec:generic_constraints}

A second family of traditional representation learning methods augments task-driven learning with explicit constraints on the latent space. These methods still optimize a task loss of the form $\mathcal{L}_{\mathrm{task}}(g_\phi(Z),T(X))$, and they additionally adapt the representation toward certain generic properties:
\begin{equation}
\label{eq:generic_constrained_objective}
    \min_{\theta,\phi}\;
    \mathbb{E}_{X,\,\tilde{X}\sim v(X)}
    \left[
        \mathcal{L}_{\mathrm{task}}
        \big(g_\phi(Z),T(X)\big)
        +\lambda\,\Omega_{\mathrm{gen}}(Z,X)
    \right],
    \qquad
    Z=f_\theta(\tilde{X}).
\end{equation}
Here $\Omega_{\mathrm{gen}}$ denotes a generic representation constraint. Such constraints may encourage compression, prior alignment, sparsity, low-energy structure, or discreteness. 

This constrained view yields a taxonomy based on the preference imposed on the latent space. Table~\ref{tab:generic_constraints} summarizes the corresponding formulations.
\textbf{Compression constraints} encourage the representation to retain only the information necessary for the target task while discarding irrelevant variability, such as information bottleneck and compact code length. 
\textbf{Distribution alignment constraints} regularize the encoder-induced latent distribution toward a prescribed prior or reference distribution. Although we only show $T(X)=Y$ as the task component in Table~\ref{tab:generic_constraints}, such constraints are not restricted to this choice and can in principle be paired with different task targets, such as reconstruction targets $T(X)=X$ or supervised targets $T(X)=Y$.
\textbf{Sparsity constraints} favor representations in which only a small number of latent coordinates are active for a given sample. 
\textbf{Energy-based constraints} assign a lower cost to latent configurations that lie in high-preference or low-energy regions. 
\textbf{Discrete or quantized constraints} restrict the representation to pass through a finite codebook. 
These constraints instantiate different choices of $\Omega_{\mathrm{gen}}$ in Eq.~\eqref{eq:generic_constrained_objective}. More details can be found in Appendix~\ref{app:generic_constraints}.

%% file: sections/4_crl.tex
\section{Causal representation learning under unified formulation}
\label{sec:crl}

\textbf{Data generation process.} Causal representation learning can be viewed as representation learning with constraints that align the learned representation with latent causal variables, mechanisms, or intervention structure. Let $Z^*=(Z_1^*,\dots,Z_D^*)$ denote latent causal variables generated by a structural causal model
\begin{equation}
\label{eq:latent_scm}
    Z_j^*
    :=
    m_j\big(Z_{\operatorname{pa}_G(j)}^*,\epsilon_j\big),
    \qquad
    j=1,\dots,D,
    \qquad
    \epsilon_i\perp \epsilon_j \;\; \text{for } i\neq j,
\end{equation}
where $G$ is a directed acyclic graph and $\operatorname{pa}_G(j)$ denotes the parents of variable $j$. The observation is generated by an unknown measurement or mixing mechanism $X=g^*(Z^*)$. The goal of CRL is to learn a representation $Z=f_\theta(\tilde{X})$ that corresponds to the latent causal variables, or at least to causally meaningful modules.

\begin{table}[tbp]
\centering
\caption{\textbf{Representative causal constraints under the unified formulation.} $U$ denotes the auxiliary information, such as the label, domain indicator, and temporal indicator, to provide sufficient observations. The pair $(v_1,v_2)$ can be the observations from different views, or with/without interventions. $d(\cdot,\cdot)$ measures discrepancy. 
In the intervention case, $\operatorname{De}_G(I)$ denotes the descendants in the causal graph $G$ under intervention $I$.
$g_\phi^{(k)}(Z_{B_k})$ denotes the sub-decoder that generates part of the observations with a latent block, and $\alpha_\phi^{(k)}(Z)$ denotes the corresponding weight, with $\sum_{k=1}^{K}\alpha_\phi^{(k)}(Z)=1$.}
\label{tab:causal_constraints}
\vspace{6pt}
\small
\setlength{\tabcolsep}{4.5pt}
\renewcommand{\arraystretch}{1.20}
\begin{tabularx}{\textwidth}{@{}L{2.45cm}Y L{2.70cm}@{}}
\toprule
\textbf{Category} & \textbf{Formulation} & \textbf{Reference} \\
\midrule

\taxcat{Sparse mechanism constraints}
&
\formtag{Notation}\quad
$\hat{E}$ denotes the estimated edge-weight structure

\vspace{2pt}
\formtag{Constraint}\quad
$
\Omega_{\mathrm{causal}}(Z,X)=\|\hat{E}\|_1
$

\vspace{2pt}
\formtag{Role}\quad
The constraint favors sparse latent mechanisms.
&
\reftight{\citep{lachapelle2022mechanism,zhang2024multiple,zheng2022sparsity,idol,el2024toward}}
\\
\softmidrule

\taxcat{Conditional-prior constraints}
&
\formtag{Notation}\quad
$p(Z\mid U)$ is the structured prior,

\vspace{2pt}
\formtag{Static constraint}\quad
$
\Omega_{\mathrm{causal}}
=
D_{\mathrm{KL}}
(q_\theta(Z\mid X)\|p(Z\mid U))
$

\vspace{2pt}
\formtag{Temporal constraint}\quad
$
\Omega_{\mathrm{causal}}
=
D_{\mathrm{KL}}
(q_\theta(Z_t\mid X_{\le t})
\|p(Z_t\mid Z_{<t},U_t))
$
&
\reftight{\citep{khemakhem2020ivae,kong2022partialdisentanglement,yaotdrl,pcl,shen2022weakly,reizinger2024cross}}
\\
\softmidrule

\taxcat{Invariance and intervention constraints}
&
\formtag{Notation}\quad
$(v_1,v_2)$ denotes pair of observations or environment;
$S^{(v_1)}_A$ is an invariant statistic on a subset $A\subseteq[D]$ of $v_1$. 
\vspace{2pt}
\formtag{Constraint}\quad
$
\Omega_{\mathrm{causal}}
=
d\!\left(
\mathcal{S}^{(v_1)}_A(Z^{(v_1)}),
\mathcal{S}^{(v_2)}_A(Z^{(v_2)})
\right)
$

\vspace{2pt}
\formtag{Intervention case}\quad
$
A=[D]\setminus \operatorname{De}_G(I)
$
&
\reftight{\citep{locatello2020weakly,yao2024unifying,ahuja2023interventional,lippe2022citris, varici2024general, varici2025score,welch2024identifiability}}
\\
\softmidrule

\taxcat{Functional constraints}
&
\formtag{Latent blocks}\quad
$
\mathcal{B}=\{B_1,\dots,B_K\},\quad
Z=(Z_{B_1},\dots,Z_{B_K})
$

\vspace{2pt}
\formtag{Additive decoder}\quad
$
g_\phi(Z)
=
\sum_{k=1}^{K}
\alpha_\phi^{(k)}(Z)\,g_\phi^{(k)}(Z_{B_k})
$
&
\reftight{\citep{locatello2020slotattention,lachapelle2023additive,ren2026causal,li2025towards}}
\\

\bottomrule
\end{tabularx}
\end{table}

\textbf{Identifiability.} A fundamental concept in CRL is identifiability, which describes how the learned representation is equivalent to the ground-truth causal factors. One commonly used notion is component-wise identifiability up to permutation and invertible scalar transformations: there exist a permutation $\pi$ and invertible scalar functions $\psi_1,\dots,\psi_D$ such that
\begin{equation}
\label{eq:component_identification}
    Z_j=\psi_j\big(Z_{\pi(j)}^*\big),
    \qquad j=1,\dots,D.
\end{equation}
A weaker but often more realistic one is block-level identifiability. Let $\mathcal{B}=\{B_1,\dots,B_K\}$ be a partition of latent coordinates. Then one may seek
\begin{equation}
\label{eq:block_identification}
    Z_{B_k}
    =
    \Psi_k\big(Z^*_{B_{\pi(k)}}\big),
    \qquad k=1,\dots,K,
\end{equation}
for invertible block-wise maps $\Psi_k$. In both cases, the remaining ambiguity must preserve enough structure for causal interpretation and intervention-aware reasoning. Additional discussion of causal representations and identifiability is provided in Appendix~\ref{app:crl_details}.

\textbf{Unified formulation.} Under the unified formulation, CRL objectives can be written as
\begin{equation}
\label{eq:crl_objective}
    \min_{\theta,\phi}\;
    \mathbb{E}_{X,\,\tilde{X}\sim v(X)}
    \left[
        \mathcal{L}_{\mathrm{task}}
        \big(g_\phi(Z),T(X)\big)
        +\lambda\,\Omega_{\mathrm{causal}}(Z,X)
    \right],
    \qquad
    Z=f_\theta(\tilde{X}).
\end{equation}
Here, the task component may be reconstruction, prediction, contrastive learning, masked prediction, or another standard representation learning objective. The defining ingredient of CRL is the causal constraint $\Omega_{\mathrm{causal}}$, which restricts the learned representation to be compatible with latent causal structure.

We summarize several categories of causal constraints in CRL. These constraints correspond to different choices of $\Omega_{\mathrm{causal}}$ in Eq.~\eqref{eq:crl_objective}.
\textbf{Sparse mechanism constraints} encode latent dependencies through an estimated edge-weight structure $\hat{E}$ and favor sparse mechanisms by penalizing $\|\hat{E}\|_1$. 
\textbf{Conditional-prior constraints} introduce a context-dependent prior $p_\eta(Z\mid U)$, where $U$ denotes an auxiliary variable, such as environment index, labels, or temporal context, and align the encoder distribution $q_\theta(Z\mid X)$ with this prior through a divergence such as $D_{\mathrm{KL}}(\cdot\|\cdot)$; in temporal settings, $Z_t$, $X_{\le t}$, and $Z_{<t}$ denote the current latent state, observed history, and past latent states. 
\textbf{Invariance and intervention constraints} use related environments or paired observations $(v_1,v_2)$ and impose agreement between invariant statistics $\mathcal{S}^{(v_1)}$ and $\mathcal{S}^{(v_2)}$. These statistics can also only be based on a subset $A\subseteq[D]$ of latent coordinates, where $[D]=\{1,\dots,D\}$, such as $\mathcal{S}^{(v_1)}_A$.
$d(\cdot,\cdot)$ denotes a discrepancy measure. For example, in the intervention case, if a latent graph $G$ is available, one may choose $A=[D]\setminus\operatorname{De}_G(I)$, where $I$ is the intervention and $\operatorname{De}_G(I)$ denotes its descendants in $G$.
\textbf{Functional constraints} impose causal inductive biases through the generative process or the structure of the latent space. These biases can take many forms~\cite{locatello2020slotattention,ren2026causal, li2025towards}. As one example, additive decoders~\cite{lachapelle2023additive} encourage a structured decomposition of the generators by partitioning the representations into blocks $\mathcal{B}=\{B_1,\dots,B_K\}$ and combining block-specific decoder branches $g_\phi^{(k)}$ with coefficients $\alpha_\phi^{(k)}(Z)$.
Table~\ref{tab:causal_constraints} summarizes the corresponding formulations. Please refer to Appendix~\ref{app:crl_details} for more details.

%% file: sections/5_mutual_benifits.tex
\section{Mutual benefits between causal and traditional representation learning}

The unified formulation presented above not only clarifies the relationship between CRL and traditional representation learning, but also suggests that the two fields can benefit each other in complementary ways. Because the two areas have developed with different emphases: CRL being more theory-driven and traditional representation learning being more application-driven. 
Each offers tools, intuitions, and evaluation perspectives that are valuable to the other. In this section, we discuss these benefits in both directions.

\subsection{A Theoretical Tool for Understanding When Structure Matters}

One important benefit of CRL to traditional representation learning is that it provides a theoretical tool for understanding when structural constraints are necessary. In many standard representation learning settings, success is defined through task performance: if a learned representation is sufficient for prediction, reconstruction, or transfer, it is often regarded as adequate. CRL sharpens this view by showing that task sufficiency does not, in general, determine the structure of the latent representation. When multiple latent organizations support the same task objective, structured constraints become important because they restrict the space of admissible solutions and favor representations aligned with stable underlying factors.



\begin{figure}[t]
    \centering
    \includegraphics[width=\linewidth]{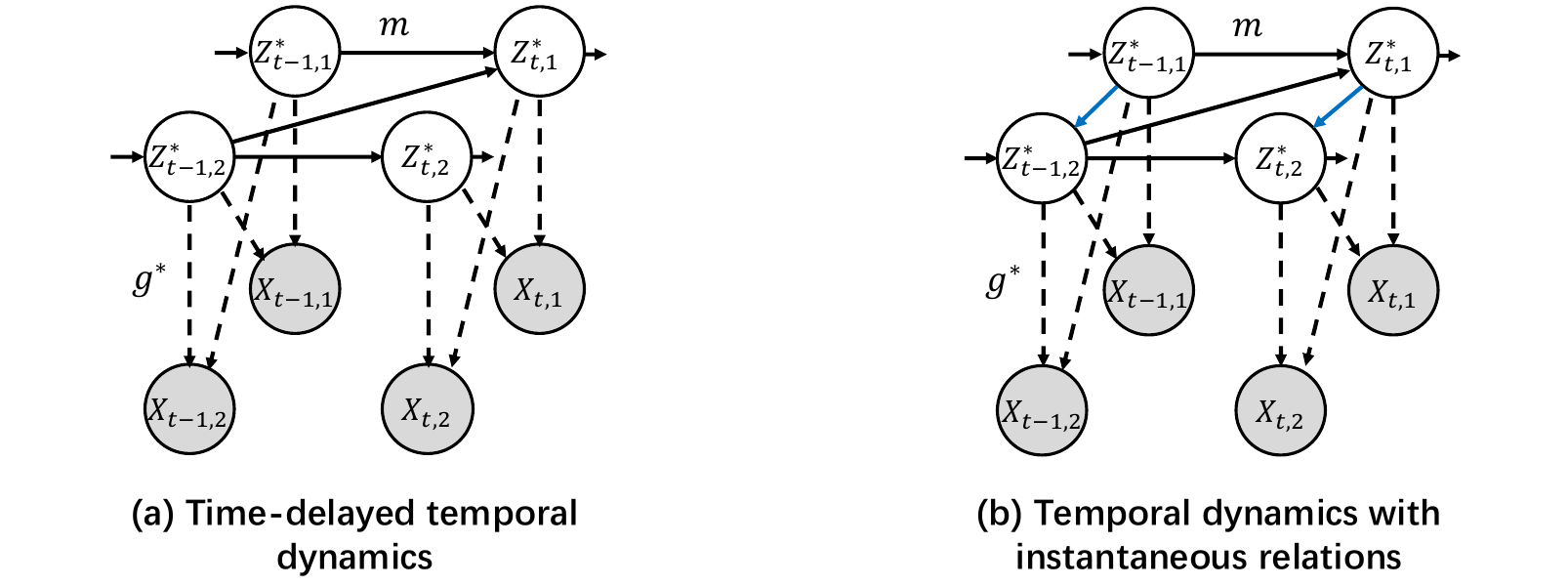}
    \caption{\textbf{Comparison of temporal dynamics with and without instantaneous relations.}
Dashed arrows denote the generation function $g*$, and solid arrows denote temporal transitions governed by $m$. 
\textbf{(a)} Without instantaneous relations, the latent variables at time $t$ depend only on variables from earlier time steps. 
\textbf{(b)} With instantaneous relations, latent variables at the same time step can additionally influence each other (blue arrows). }
\label{fig:temporal_instantaneous}
\vspace{-0.3cm}
\end{figure}

\textbf{A video prediction example.}
We use video prediction as a simple example to illustrate when structured constraints become necessary. Let $X_t$ denote the observed video frame at time $t$, and let
$Z_t^*=(Z_{t,1}^*,\ldots,Z_{t,D}^*)$ denote the ground-truth latent causal variables that generate the frame. As shown in Figure~\ref{fig:temporal_instantaneous}(a), the generation process without instantaneous relations is given by
\begin{equation}
    X_t = g^*(Z_t^*),~~~~~~  Z_{t,j}^*=m_j\big(Z_{<t,\mathrm{pa}_G(j)}^*, \epsilon_{t,j}\big),
\end{equation}
where $\epsilon_{t,j}$ is the independent noise. 

A standard video prediction objective learns a representation from past observations,
\[
    Z_t=f_\theta(X_{\leq t-1}),
\]
and uses it to predict the next frame $X_t$. This task component encourages $Z_t$ to preserve information from the past that is useful for prediction.
Here, the parents of $Z_{t,i}^*$ come only from previous time steps. 
If all relevant latent variables contribute to the prediction of $X_t$, then the video prediction objective can provide direct learning signal for these variables. Dropping an important component of $Z_t^*$ would generally reduce predictive information about $X_t$. In this case, the task objective may be sufficient for learning a representation that supports the prediction task.

The situation changes when instantaneous relations exist among the latent variables. In this case, as illustrated in part (b) of Figure~\ref{fig:temporal_instantaneous}, a latent variable at time $t$ may depend not only on previous latent variables, but also on other latent variables at the same time step:
\begin{equation}
    Z_{t,j}^*
    :=
    m_j\big(Z_{<t,\mathrm{pa}_G^d(j)}^*, Z_{t,\mathrm{pa}_G^e(j)}^*, \epsilon_{t,j}\big),
    \qquad j=1,\ldots,D,
\end{equation}
where $\mathrm{pa}_G^d(j)$ denotes time-delayed parents and $\mathrm{pa}_G^e(j)$ denotes instantaneous parents. For example, in a video of human motion, the latent state of one joint may instantaneously constrain the state of another joint within the same frame.

In this setting, the video prediction objective alone may be insufficient for recovering the latent organization. A representation can be useful for predicting $X_t$ without separating the components of $Z_t^*$ or recovering the instantaneous relations among them. For example, it may encode an entangled feature that supports accurate frame prediction but does not correspond to the causal variables. Thus, structured constraints such as sparsity of latent influences are needed precisely in this underdetermined case to make the prediction under intervention. They restrict the admissible learned representations and help learn the representations whose components and dependencies are compatible with the ground-truth latent causal process.

\subsection{How Traditional Representation Learning Helps CRL in Practice}

Traditional representation learning can benefit CRL by providing practical guidance on how representations should be learned and optimized. While CRL research has focused mainly on structural constraints, traditional representation learning has developed a rich set of task objectives and empirical design principles. These practical insights are especially valuable for CRL because the effect of a structured constraint can vary substantially depending on the task objective with which it is paired.

Traditional representation learning offers many task components, including reconstruction, classification, contrastive learning, masked prediction, autoregressive prediction, and transformation prediction. Each task encourages the representation to preserve different information. For example, a reconstruction objective encourages the representation to retain detailed information about the input, while a contrastive objective emphasizes information shared across augmented views. A masked prediction objective encourages contextual and structural completion. When these tasks are combined with the same causal constraint, they may lead to different latent representations. Thus, traditional representation learning helps CRL move beyond treating the task objective as fixed, and instead encourages a more systematic study of which task should be paired with which causal constraint.

%% file: sections/6_experiments.tex
\section{Experiments}
\label{sec:exp}

In this section, we evaluate how the effectiveness of causal constraints depends on the task objectives with which they are paired. We first describe the experimental setup in Section~\ref{subsec:exp_setup}, and then present and discuss the results in Section~\ref{subsec:exp_res}.

\subsection{Experimental setup}
\label{subsec:exp_setup}

\textbf{Dataset.}
Since our goal is to recover the true latent variables, we require datasets with ground-truth causal factors. Such annotations are difficult to obtain in real-world datasets. Therefore, we conduct our experiments on the CausalVerse benchmark~\citep{causalverse}, a high-quality synthetic dataset that provides full access to ground-truth latent variables.
CausalVerse contains 200,000 images and 3 million video frames across 24 sub-scenes spanning four domains. 
This controlled setting enables precise evaluation of whether learned representations align with the true causal factors.
We focus on two representative and challenging scenarios: image-based and video-based settings, which reflect common real-world applications. For the video setting, we use the \textit{Robotics Study}, which contains 57-dimensional dynamic latent variables. For the image setting, we use \textit{Ball on the Slope}, which contains 7-dimensional latent variables.

\textbf{Backbone and Task Components.}
For the video setting, we adopt TDRL~\citep{yaotdrl} as the base method. TDRL combines a reconstruction-based task objective with conditional-prior constraints. To isolate the role of the task objective, we keep the conditional-prior constraints fixed and vary only the task component. Building on the default \texttt{reconstruction} objective (frame-wise reconstruction), we replace it with several commonly used task components in video-based CRL methods.
Specifically, \texttt{Contrastive Learning} adopts a CPC-style objective that matches the current latent history with the correct future feature among in-batch candidates. \texttt{Next-Frame Prediction} trains the model to predict the next frame from the current latent state, forming a temporal target-prediction task. \texttt{Mid-Latent Reconstruction} shifts the reconstruction target from the observation space to an intermediate latent representation. \texttt{Prototype-based Learning} clusters the learned latent representation $z$ and applies a classification loss over the resulting assignments. \texttt{Masked Reconstruction} trains the model to recover missing content from partially observed (masked) inputs.

For the image setting, we adopt iMSDA~\citep{kong2022partialdisentanglement} as the base method, whose unsupervised component similarly integrates a reconstruction-based objective with conditional-prior constraints. Extending beyond the default \texttt{reconstruction} objective, we incorporate several representative task components commonly used in image-based CRL methods.
In particular, \texttt{Denoising Reconstruction} uses a corrupted input while reconstructing the clean target, remaining within the reconstruction family. \texttt{Masked Reconstruction} again corresponds to recovering missing information from partial observations. \texttt{Cross-view Feature Prediction} uses one augmented view to predict the feature of another view, forming a target-prediction objective. \texttt{Contrastive Learning} enforces agreement between positive pairs while separating negative samples. \texttt{Prototype-based Learning} clusters latent representations and applies a classification loss over prototype assignments.

Please refer to Table~\ref{tab:task_driven_rl} for the formulation of each task component; further implementation details are provided in Appendix~\ref{section:experiment_details}. We also provide an additional study in Appendix~\ref{subsec:sparsity_task_variants}, where we examine whether the choice of task objective continues to matter under sparsity constraints.

\textbf{Metrics.}
To evaluate how well the learned representations recover the true latent variables, we use two standard metrics: the mean correlation coefficient (MCC) and the coefficient of determination ($R^2$). These metrics quantify the alignment between learned representations and ground-truth latent factors, with higher values indicating stronger alignment with the underlying causal factors.

\begin{table}[t]
\centering
\caption{Performance of different task objectives when paired with the same causal constraint mechanism in video and image settings.}
\begin{subtable}[t]{0.48\textwidth}
\centering
\caption{Task objective comparison with fixed causal constraints in the video setting.}
\label{tab:video_task_comparison}
\begin{tabular}{l c c}
\hline
\textbf{Task} & \textbf{MCC} & $\boldsymbol{R^2}$ \\
\hline
Reconstruction (TDRL) & 0.25 & 0.82 \\
Contrastive Learning & 0.36 & 0.92 \\
Next-Frame Prediction & 0.24 & 0.82 \\
Mid-Latent Reconstruction & 0.08 & 0.69 \\
Prototype-based Learning & 0.16 & 0.80 \\
Masked Reconstruction & 0.08 & 0.72 \\
\hline
\end{tabular}
\end{subtable}
\hfill
\begin{subtable}[t]{0.48\textwidth}
\centering
\caption{Task objective comparison with fixed causal constraints in the image setting.}
\label{tab:image_task_comparison}
\begin{tabular}{l c c}
\hline
\textbf{Task} & \textbf{MCC} & $\boldsymbol{R^2}$ \\
\hline
Reconstruction (iMSDA) & 0.48 & 0.91 \\
Contrastive Learning & 0.62 & 0.94 \\
Denoising Reconstruction & 0.57 & 0.93 \\

Cross-view Prediction & 0.44 & 0.91 \\

Prototype-based Learning & 0.48 & 0.88 \\
Masked Reconstruction & 0.38 & 0.88 \\
\hline
\end{tabular}
\end{subtable}

\vspace{-0.3cm}
\label{tab:task_comparison_side_by_side}
\end{table}




\subsection{Results and discussion}
\label{subsec:exp_res}

To understand the effect of task component, we compare performance across different objectives in both video and image settings, as shown in Table~\ref{tab:task_comparison_side_by_side}.

\textbf{Task objectives critically modulate the effect of causal constraints.}
A clear trend emerges: changing the task objective while keeping the causal constraints fixed leads to substantial variation in performance. This suggests that causal constraints alone are not sufficient to guarantee recovery of the underlying latent structure. 
In both settings, contrastive learning consistently achieves the best performance, indicating that objectives emphasizing relational or discriminative structure may better align with causal factorization. In contrast, objectives such as \texttt{mid-latent reconstruction} and \texttt{masked reconstruction} (in the video setting) perform significantly worse, despite being paired with the same constraints. This highlights that not all reconstruction-style objectives are equally compatible with causal structure learning.

\textbf{Differences between image and video settings.}
While the overall trend is consistent across both modalities, we observe that the impact of task choice is more pronounced in the video setting. This may be due to the higher dimensionality and temporal dependencies of the latent variables, which require stronger inductive biases from the task objective. In contrast, the image setting shows stable performance across tasks, while contrastive and denoising objectives still provide clear gains.

\textbf{Implications.}
These findings challenge the common assumption that designing better causal constraints alone is sufficient for improving CRL. Instead, they emphasize the need to jointly design task objectives and constraint mechanisms. A well-matched task–constraint pair can significantly enhance representation quality, while a mismatch can severely limit the benefits of causal structure, even when the constraints are correctly specified.

%% file: sections/7_conclusion.tex
\section{Conclusion}

We studied the relationship between CRL and traditional representation learning from a learning perspective. We argued that, although the two fields have developed with different emphases, they can be understood within a unified framework consisting of a task component and a constraint component. This view clarifies that CRL is distinguished not only by the use of structured constraints, but also by the settings in which such constraints become necessary beyond standard task objectives. It also shows that the two fields can mutually benefit each other: CRL contributes theoretical tools for understanding when structure matters, while traditional representation learning offers practical insights on task design and objective choice that can improve CRL in real applications. Our analysis and experiments further suggest that the practical role of causal constraints cannot be understood in isolation, but depends critically on the task with which they are paired. We hope this perspective helps bridge the gap between the two fields and encourages closer interaction between theory-driven and application-driven research on learned representations.  \textbf{Limitation and future work:} A limitation of the present work is that our analysis of the role of task components is mainly grounded in empirical observations. Developing a more principled account of how to choose appropriate task--constraint combinations remains an important direction for future work. 

%% file: appendix/1-notation.tex
\section{Notation and Terminology}
We summarize the notations used throughout the paper in Table~\ref{tab:notation_terminology}.

\begingroup
\small
\setlength{\tabcolsep}{5pt}
\renewcommand{\arraystretch}{1.13}

\begin{longtable}{@{}L{2.20cm}L{6.95cm}L{4.95cm}@{}}
\toprule
\textbf{Notation / Term} & \textbf{Explanation} & \textbf{Support / Range} \\
\midrule
\endfirsthead

\toprule
\textbf{Notation / Term} & \textbf{Explanation} & \textbf{Support / Range} \\
\midrule
\endhead

\midrule
\multicolumn{3}{r}{\footnotesize\emph{Continued on next page}}\\
\endfoot

\bottomrule
\\[-6pt]
\caption{Notation and terminology used throughout the paper.}
\label{tab:notation_terminology}
\endlastfoot

\firstnotecat{Index}

$i,j,k,m$
&
Generic indices for samples, variables, classes, codebook entries, decoder branches, or latent blocks.
&
Problem-dependent finite index sets.
\\

$c$
&
Class index.
&
$c\in\{1,\dots,C\}$.
\\

$t$
&
Time index.
&
$t\in\mathbb{N}^{+}$.
\\

$B$
&
Mini-batch size.
&
$B\in\mathbb{N}^{+}$.
\\

$C$
&
Number of classes.
&
$C\in\mathbb{N}^{+}$.
\\

$K$
&
Number of transformations, clusters, prototypes, codebook entries, or latent blocks.
&
$K\in\mathbb{N}^{+}$; meaning is context-dependent.
\\

$D$
&
Number of ground-truth latent causal variables or latent coordinates in CRL.
&
$D\in\mathbb{N}^{+}$.
\\

\notecat{Variable}

$X$
&
Observed sample.
&
$X\in\mathcal{X}$, drawn from $p_{\mathrm{data}}(X)$.
\\

$\mathcal{X}$
&
Observation space.
&
Problem-dependent.
\\

$p_{\mathrm{data}}(X)$
&
Data distribution.
&
Probability density or mass function over $\mathcal{X}$.
\\

$\tilde{X}$
&
Observed view given to the learner.
&
$\tilde{X}\sim\mathcal{V}(X)$.
\\

$T(X)$
&
Target signal induced by the task.
&
Problem-dependent; e.g., $Y$, $X$, $X_M$, $v_2(X)$, or $q(X)$.
\\

$Y$
&
Supervised target or label.
&
$Y\in\{1,\dots,C\}$ for classification.
\\

$Z$
&
Learned latent representation.
&
$Z=f_\theta(\tilde{X})\in\mathcal{Z}$.
\\

$\mathcal{Z}$
&
Latent representation space.
&
Often $\mathcal{Z}\subseteq\mathbb{R}^{D}$ or a structured latent space.
\\

$Z^*$
&
Ground-truth latent causal variables.
&
$Z^*=(Z_1^*,\dots,Z_D^*)\in\mathcal{Z}^*$.
\\

$Z_j^*$
&
The $j$-th ground-truth latent causal variable.
&
Scalar or vector-valued, depending on the latent SCM.
\\

$Z_t$
&
Learned latent state at time $t$.
&
Used in temporal prediction and temporal CRL.
\\

$Z_t^*$
&
Ground-truth latent causal variables at time $t$.
&
$Z_t^*=(Z_{t,1}^*,\dots,Z_{t,D}^*)$.
\\

$Z_{t,j}^*$
&
The $j$-th ground-truth latent causal variable at time $t$.
&
Problem-dependent.
\\

$X_t$
&
Observed video frame at time $t$.
&
$X_t\in\mathcal{X}$.
\\

$X_{\leq t-1}$
&
Past observations up to time $t-1$.
&
Sequence prefix used in video prediction.
\\

$X_{\leq t},X_{>t}$
&
Prefix and future segment of a sequence.
&
Used in autoregressive or temporal prediction.
\\

$X_M,X_{M^{\urcorner}}$
&
Masked and visible parts of the input.
&
$M$ and $M^{\urcorner}$ partition observed positions.
\\

$X_i,\hat{X}_i$
&
Input element or sample and its prediction.
&
Token, pixel, patch, feature element, or sample indexed by $i$.
\\

$\epsilon$
&
Exogenous noise.
&
$\epsilon_i\perp\epsilon_j$ for $i\neq j$ in the latent SCM.
\\

$U$
&
Auxiliary information for conditional-prior constraints.
&
May denote a label, domain indicator, temporal indicator, environment index, or other auxiliary variable.
\\

$z_e$
&
Continuous encoder output before quantization.
&
$z_e=f_\theta(X)$.
\\

$e_k$
&
The $k$-th code vector.
&
$e_k\in\{e_1,\dots,e_K\}$.
\\

\notecat{Set / Graph / Structural Object}

$M,M^{\urcorner}$
&
Masked and visible positions.
&
$M\cup M^{\urcorner}$ equals the set of observed positions.
\\

$\mathcal{N}^-$
&
Negative candidate set for sample $X$.
&
Used in contrastive learning.
\\

$\mathcal{N}_i^-$
&
Negative candidate set for sample $X_i$.
&
Used in sample-indexed contrastive objectives.
\\

$\{e_k\}_{k=1}^{K}$
&
Finite codebook.
&
Used in discrete or quantized constraints.
\\

$G$
&
Directed acyclic graph.
&
Directed acyclic graph over latent causal variables.
\\

$\operatorname{pa}_G(j)$
&
Parents of node $j$ in $G$.
&
Subset of latent variables.
\\

$\operatorname{De}_G(I)$ 
&
Descendants of intervention target $I$ in $G$.
&
Subset of latent variables or modules.
\\

$\hat{E}$
&
Estimated edge-weight structure.
&
Used to encode latent dependencies in sparse mechanism constraints.
\\

$\mathcal{B}$
&
Partition of latent coordinates.
&
$\mathcal{B}=\{B_1,\dots,B_K\}$.
\\

$B_k$
&
The $k$-th latent block.
&
$B_k\subseteq\{1,\dots,D\}$ and $B_i\cap B_j=\emptyset$ for $i\neq j$.
\\

$A$
&
Invariant subset of latent coordinates.
&
$A\subseteq[D]$, where $[D]=\{1,\dots,D\}$.
\\

$I$
&
Intervention target.
&
$I\subseteq[D]$ or a subset of latent modules.
\\

\notecat{Function / Map}

$v$
&
View-generation process.
&
$\tilde{X}\sim v(X)$.
\\

$f_\theta$
&
Encoder.
&
Maps an observed view $\tilde{X}$ to a latent representation $Z$.
\\

$g_\phi$
&
Predictor, decoder, projection head, or task head.
&
Maps $Z$ to a task-dependent output space.
\\

$g^*$
&
Ground-truth measurement or mixing mechanism.
&
Used in causal data generation process.
\\

$m_j$
&
Structural mechanism for $Z_j^*$.
&
$Z_j^*:=m_j(Z_{\operatorname{pa}_G(j)}^*,\epsilon_j)$.
\\

$v^{\mathrm{cor}}(X)$
&
Corrupted version of $X$.
&
Used in denoising reconstruction.
\\


$v^{\mathrm{rot}}$
&
Rotation transformation.
&
Special case of $v$ in transformation correction.
\\

$v^{\mathrm{jig}}$
&
Jigsaw or patch-permutation transformation.
&
Special case of $v$ in transformation correction.
\\

$v_1(\cdot),v_2(\cdot)$
&
Two related views, observations, or environments.
&
Used in contrastive learning and invariance/intervention constraints.
\\

$h,h^+$
&
Projected positive-pair representations.
&
$h=g_\phi(f_\theta(v_1(X)))$, $h^+=g_\phi(f_\theta(v_2(X)))$.
\\

$h_i,h_i^+$
&
Projected positive-pair representations for sample $X_i$.
&
$h_i=g_\phi(f_\theta(v_1(X_i)))$, $h_i^+=g_\phi(f_\theta(v_2(X_i)))$.
\\

$s(\cdot,\cdot)$
&
Similarity function.
&
Used in contrastive objectives, e.g., cosine similarity.
\\

$q(X)$
&
Cluster or prototype assignment.
&
$q(X)\in\{1,\dots,K\}$.
\\

$q_\theta(Z\mid X)$
&
Stochastic encoder or approximate posterior.
&
Distribution over latent representation $Z$ given $X$.
\\

$p_\phi(Y\mid Z)$
&
Predictive target distribution.
&
Used in target prediction or supervised objectives.
\\

$p_\phi(v\mid Z)$
&
Predictive distribution over transformations.
&
Used in transformation correction.
\\

$p_\phi(q(X)\mid Z)$
&
Predictive distribution over cluster or prototype assignments.
&
Used in cluster or prototype learning.
\\

$p_\phi(X\mid Z)$
&
Decoder likelihood.
&
Used in reconstruction or variational reconstruction objectives.
\\

$p(Z)$
&
Reference prior.
&
Often a simple prior, e.g., standard Gaussian.
\\

$p_\psi(Z)$
&
Energy-based latent prior.
&
$p_\psi(Z)\propto\exp(-E_\psi(Z))$.
\\

$E_\psi(Z)$
&
Energy function.
&
Defines an energy-based latent constraint.
\\

$p(Z\mid U)$
&
Structured prior conditioned on auxiliary information.
&
Used in conditional-prior constraints.
\\

$p_\eta(Z\mid U)$
&
Parametric structured prior conditioned on auxiliary information.
&
Often used to instantiate $p(Z\mid U)$.
\\

$\psi_j$
&
Component-wise invertible map.
&
Used in $Z_j=\psi_j(Z_{\pi(j)}^*)$.
\\

$\Psi_k$
&
Block-wise invertible map.
&
Used in $Z_{B_k}=\Psi_k(Z^*_{B_{\pi(k)}})$.
\\

$\mathcal{S}^{(v_1)}_{A}$
&
Invariant statistic.
&
Computed on subset $A$ for the view, observation, or environment $v_1$.
\\

$\mathcal{S}^{(v_2)}_{A}$
&
Invariant statistic.
&
Computed on subset $A$ for the view, observation, or environment $v_2$.
\\

$d(\cdot,\cdot)$
&
Discrepancy measure.
&
Used to compare invariant statistics.
\\

$\alpha_\phi^{(k)}(Z)$
&
Combination coefficient for decoder branch $k$.
&
Used in additive functional constraints, often with $\sum_{k=1}^{K}\alpha_\phi^{(k)}(Z)=1$.
\\

$g_\phi^{(k)}$
&
Block-specific decoder branch.
&
Depends only on latent block $Z_{B_k}$.
\\

\notecat{Objective / Constraint}

$\mathcal{L}_{\mathrm{task}}$
&
Task loss.
&
Measures prediction or reconstruction error between $g_\phi(Z)$ and $T(X)$.
\\

$\Omega$
&
Additional constraint term in the unified objective.
&
Specified by the constraint component.
\\

$\Omega_{\mathrm{gen}}$
&
Generic non-causal constraint.
&
Used for compression, prior alignment, sparsity, energy-based preferences, or quantization.
\\

$\Omega_{\mathrm{causal}}$
&
Causal constraint.
&
Encodes assumptions about latent variables, mechanisms, interventions, or modularity.
\\

$\lambda$
&
Constraint weight.
&
$\lambda\geq 0$.
\\

$\beta$
&
Regularization strength.
&
Usually $\beta>0$.
\\

$I(X;Z)$
&
Mutual information.
&
Used in compression constraints.
\\

$D_{\mathrm{KL}}(\cdot\|\cdot)$
&
Kullback--Leibler divergence.
&
Used in variational, bottleneck, sparsity, and prior-alignment objectives.
\\

$\|Z\|_1$
&
Sparsity penalty on the representation.
&
Used in sparse representation learning.
\\

$\|\hat{E}\|_1$
&
Sparsity penalty on the estimated edge-weight structure.
&
Used in sparse mechanism constraints.
\\

$\rho$
&
Target sparsity level.
&
Typically $\rho\in(0,1)$.
\\

$\hat{\rho}$
&
Empirical activation rate.
&
Used in activation sparsity penalties.
\\

$\tau$
&
Temperature parameter.
&
$\tau>0$.
\\

$\mathrm{sg}[\cdot]$
&
Stop-gradient operator.
&
Used in vector-quantized objectives.
\\

\notecat{Operator / Transformation}

$\arg\min$
&
Minimizer operator.
&
Used for nearest-codebook assignment.
\\

$\operatorname{softmax}(\cdot)$
&
Categorical normalization map.
&
Used in classification and prototype prediction.
\\

$\perp$
&
Statistical independence.
&
Used for exogenous noises.
\\

$\pi$
&
Permutation.
&
Used in identifiability and block-level recovery.
\\

$\operatorname{pa}_G^d(j)$
&
Time-delayed parents of variable $j$.
&
Used in temporal examples with delayed causal relations.
\\

$\operatorname{pa}_G^e(j)$
&
Instantaneous parents of variable $j$.
&
Used in temporal examples with instantaneous causal relations.
\\

$Z_{<t,\operatorname{pa}_G(j)}^*$
&
Previous-time parents of $Z_{t,j}^*$.
&
Used in the video prediction example.
\\

$Z_{<t,\operatorname{pa}_G^d(j)}^*$
&
Time-delayed parent variables of $Z_{t,j}^*$.
&
Used in the instantaneous-relation temporal example.
\\

$Z_{t,\operatorname{pa}_G^e(j)}^*$
&
Instantaneous parent variables of $Z_{t,j}^*$.
&
Used in the instantaneous-relation temporal example.
\\

\end{longtable}

\endgroup

%% file: appendix/2-additional-details.tex
\section{Additional details on representative representation learning methods}
\label{app:representative_methods}

This appendix expands the representative methods and constraints summarized in Sections~\ref{sec:traditional_rl} and~\ref{sec:crl}. 
The goal is not to provide a comprehensive survey, but to make explicit how common objectives and constraints instantiate the unified formulation.

\subsection{Traditional representation learning: task-driven methods}
\label{app:task_driven_rl}

When the representation constraint is absent, or only plays a weak regularizing role, the unified formulation reduces to
\begin{equation}
\label{eq:app_task_driven_objective}
    \min_{\theta,\phi}\;
    \mathbb{E}_{X,\tilde{X}\sim v(X)}
    \Big[
        \mathcal{L}_{\mathrm{task}}
        \big(g_\phi(f_\theta(\tilde{X})),\,T(X)\big)
    \Big].
\end{equation}
Here $\tilde{X}$ denotes the observed view generated from the underlying sample $X$, $T(X)$ denotes the target signal, and $Z=f_\theta(\tilde{X})$ denotes the learned representation. Under this view, traditional representation learning methods are distinguished primarily by two choices: the view-generation process $\mathcal{V}$, which determines what the learner observes, and the target signal $T(X)$, which determines what information the representation must preserve. In most cases, $Z$ is optimized to be sufficient for predicting a designated label, view, transformation, or assignment of the underlying sample~\citep{bengio2013representation}.

\subsubsection{Target prediction}

Target prediction is the most direct instantiation of the framework. The learner observes the full input and predicts an externally provided target:
\begin{equation}
    \tilde{X}=X,\qquad T(X)=Y,\qquad Z=f_\theta(X).
\end{equation}
When $Y$ is categorical, this reduces to supervised classification. A canonical task loss is
\begin{equation}
    \mathcal{L}_{\mathrm{task}}
    =
    -\log p_\phi(Y\mid Z).
\end{equation}
Equivalently, if $Y$ is represented as a class label in $\{1,\dots,C\}$, then $p_\phi(Y\mid Z)$ may be implemented by a softmax classifier. The resulting objective is
\begin{equation}
    \min_{\theta,\phi}\;
    \mathbb{E}_{(X,Y)}
    \Big[
        -\log p_\phi(Y\mid f_\theta(X))
    \Big].
\end{equation}
Under this formulation, the learned representation is explicitly encouraged to retain information relevant for label prediction, and is therefore naturally task-dependent~\citep{lecun1998gradient,krizhevsky2012imagenet,he2016resnet,yosinski2014transfer}.

\subsubsection{Reconstruction}

Reconstruction uses the observation itself as the target signal:
\begin{equation}
    \tilde{X}=X,\qquad T(X)=X,\qquad Z=f_\theta(X).
\end{equation}
For continuous observations, a standard choice is the squared reconstruction error
\begin{equation}
    \mathcal{L}_{\mathrm{task}}
    =
    \|X-g_\phi(Z)\|_2^2.
\end{equation}
The learning objective becomes
\begin{equation}
    \min_{\theta,\phi}\;
    \mathbb{E}_{X}
    \Big[
        \|X-g_\phi(f_\theta(X))\|_2^2
    \Big].
\end{equation}
The representation is thus trained to preserve information sufficient for reconstructing the original input, so that $Z$ acts as a compressed code of the observation. A closely related variant is denoising reconstruction, where the encoder takes a corrupted view $\tilde{X}=v^{\mathrm{cor}}(X)$ as input but is still trained to recover the clean sample $X$. This changes the view-generation process while keeping the same reconstruction target. Classical autoencoders, denoising autoencoders, and stacked denoising autoencoders are canonical examples of this family~\citep{hinton2006reducing,vincent2008extracting,vincent2010stacked,masci2011stacked}.

\subsubsection{Partial information prediction}

Partial information prediction covers methods in which the model observes only part of the original sample and predicts the missing part. Let $M$ denote the set of masked positions and $M^{\urcorner}$ the visible positions. Then
\begin{equation}
    \tilde{X}=X_{M^{\urcorner}},\qquad T(X)=X_M,\qquad Z=f_\theta(X_{M^{\urcorner}}).
\end{equation}
For continuous inputs or image patches, a common task loss is masked reconstruction:
\begin{equation}
    \mathcal{L}_{\mathrm{task}}^{\mathrm{mask}}
    =
    \|X_M-g_\phi(Z)\|_2^2.
\end{equation}
For discrete tokens, the corresponding loss is usually a negative log-likelihood:
\begin{equation}
    \mathcal{L}_{\mathrm{task}}^{\mathrm{mask}}
    =
    -\log p_\phi(X_M\mid Z).
\end{equation}
Accordingly, the optimization problem can be written as
\begin{equation}
    \min_{\theta,\phi}\;
    \mathbb{E}_{X,M}
    \Big[
        \|X_M-g_\phi(f_\theta(X_{M^{\urcorner}}))\|_2^2
    \Big],
\end{equation}
or, in discrete settings, by replacing the squared loss with the corresponding negative log-likelihood. Such methods encourage the representation to capture contextual and structural regularities that support the recovery of missing content~\citep{devlin2019bert,he2022mae,pathak2016context}.

Autoregressive prediction can be viewed as a temporal instance of partial information prediction, where the observed view is a prefix or past segment of the data:
\begin{equation}
    \tilde{X}=X_{\le t},\qquad T(X)=X_{>t}.
\end{equation}
A canonical task loss is the autoregressive negative log-likelihood
\begin{equation}
    \mathcal{L}_{\mathrm{task}}^{\mathrm{AR}}
    =
    -\log p_\phi(X_{>t}\mid X_{\le t}),
\end{equation}
or, equivalently, a product of conditional likelihoods over future positions. Under this formulation, the learned representation must encode sequential dependencies that are predictive of future observations~\citep{bengio2003neural,graves2013generating,oord2016wavenet,oord2018cpc}.

\subsubsection{Transformation correction}

Transformation correction constructs an artificial target from the transformation applied to the input. Let $v$ denote a transformation selected from a finite family of transformations. Then
\begin{equation}
    \tilde{X}=v(X),\qquad T(X)=v,\qquad Z=f_\theta(v(X)).
\end{equation}
A canonical task loss is
\begin{equation}
    \mathcal{L}_{\mathrm{task}}
    =
    -\log p_\phi(v\mid Z).
\end{equation}
The learning problem becomes
\begin{equation}
    \min_{\theta,\phi}\;
    \mathbb{E}_{X,v}
    \Big[
        -\log p_\phi(v\mid f_\theta(v(X)))
    \Big].
\end{equation}
The resulting representation is encouraged to preserve semantic and geometric cues that make the applied transformation identifiable.

Rotation prediction is a representative example, where $v=v^{\mathrm{rot}}$ is a discrete rotation~\citep{gidaris2018rotation}. Jigsaw prediction is another example, where $v=v^{\mathrm{jig}}$ is a patch permutation and the model predicts the permutation index rather than reconstructing the original image~\citep{noroozi2016jigsaw}. More broadly, many pretext-task-based methods differ in how they define the surrogate target $T(X)$, for example via context prediction, inpainting, or colorization~\citep{doersch2015context,pathak2016context,zhang2016colorization}.

\subsubsection{Contrastive learning}

Contrastive learning may appear different from standard prediction-based objectives because it often does not reconstruct an explicit target in the raw input space. Nevertheless, it fits into the same framework if the target signal is interpreted as another semantically related view of the same underlying sample.

Let $v_1(X)$ and $v_2(X)$ denote two views of the same sample, for example obtained by two stochastic augmentations. One may write
\begin{equation}
    \tilde{X}=v_1(X),\qquad T(X)=v_2(X).
\end{equation}
For contrastive methods, let
\begin{equation}
    h_i=g_\phi(f_\theta(v_1(X_i))),\qquad
    h_i^{+}=g_\phi(f_\theta(v_2(X_i))).
\end{equation}
A canonical task loss is InfoNCE:
\begin{equation}
    \mathcal{L}_{\mathrm{task}}^{\mathrm{InfoNCE}}
    =
    -\log
    \frac{\exp(s(h_i,h_i^+)/\tau)}
    {\exp(s(h_i,h_i^+)/\tau)+
    \sum_{H\in\mathcal{N}_i^-}\exp(s(h_i,H)/\tau)},
\end{equation}
where $s(\cdot,\cdot)$ is a similarity function, $\tau$ is a temperature parameter, and $\mathcal{N}_i^-$ denotes the set of negative candidates for sample $X_i$. The full objective averages this loss over samples. Under this formulation, the representation is encouraged to preserve information shared across related views while discarding view-specific nuisance variation~\citep{oord2018cpc,he2020moco,chen2020simclr,grill2020bootstrap,chen2021exploring,zbontar2021barlow}.

\subsubsection{Cluster / prototype learning}

Cluster or prototype learning constructs the target signal from the organization of samples in representation space. Rather than predicting a manually specified label or a predefined transformation, these methods learn by matching each sample to a cluster or prototype induced from the data itself. Let $q(X)\in\{1,\dots,K\}$ denote a cluster or prototype assignment obtained from the current representation space. Consistent with the main text, we write the basic form as
\begin{equation}
    \tilde{X}=X,\qquad T(X)=q(X),\qquad Z=f_\theta(X).
\end{equation}
A canonical task loss is
\begin{equation}
    \mathcal{L}_{\mathrm{task}}^{\mathrm{cluster}}
    =
    -\log p_\phi(q(X)\mid Z).
\end{equation}
The resulting objective is
\begin{equation}
    \min_{\theta,\phi}\;
    \mathbb{E}_{X}
    \Big[
        -\log p_\phi(q(X)\mid f_\theta(X))
    \Big].
\end{equation}
The assignments are periodically updated from the current features, for example by clustering the representation space or by matching features to a learned set of prototypes. In this way, the representation is encouraged to preserve structure that supports grouping samples into semantically coherent partitions. Variants may replace the identity view with an augmented view $v(X)$ or enforce consistency between assignments induced by two views. Under this formulation, the learned representation is driven to capture stable group-level structure shared across views while discarding nuisance variation that is not predictive of the induced assignment~\citep{xie2016unsupervised,caron2018deepcluster,caron2020unsupervised,ji2019invariant}.

\subsection{Traditional representation learning: generic constraints}
\label{app:generic_constraints}

The unified formulation also covers methods in which representation learning is shaped not only by the target signal, but also by explicit constraints on the learned representation. In these methods, the latent variable is still optimized to predict $T(X)$ from an observed view $\tilde{X}$, but the optimization is additionally biased toward representations with particular structural properties:
\begin{equation}
\label{eq:app_generic_constraint_objective}
    Z=f_\theta(\tilde{X}),\qquad
    \min_{\theta,\phi}\;
    \mathbb{E}_{X,\tilde{X}\sim v(X)}
    \Big[
        \mathcal{L}_{\mathrm{task}}\big(g_\phi(Z),\,T(X)\big)
        + \lambda\,\Omega_{\mathrm{gen}}(Z,X)
    \Big].
\end{equation}
Unlike purely task-driven methods, the methods considered here also differ in how they constrain the geometry, information content, or combinatorial structure of the latent space. These additional constraints do not replace target-driven learning; rather, they refine it by favoring representations that are compressed, prior-aligned, sparse, low-energy, or discrete.

\subsubsection{Compression constraints}

Compression constraints encourage the representation to retain only the information necessary for the target task. The guiding idea is that a useful representation should discard irrelevant variability while preserving target-relevant content.

A representative example is the information bottleneck principle. In this setting,
\begin{equation}
    \tilde{X}=X,\qquad T(X)=Y,\qquad Z\sim q_\theta(Z\mid X).
\end{equation}
A canonical task loss is the negative log-likelihood of the label given the representation:
\begin{equation}
    \mathcal{L}_{\mathrm{task}}
    =
    -\log p_\phi(Y\mid Z).
\end{equation}
The ideal information-bottleneck objective can be written as
\begin{equation}
    \min_{\theta,\phi}\;
    \mathbb{E}_{(X,Y)}\mathbb{E}_{Z\sim q_\theta(Z\mid X)}
    \big[
        -\log p_\phi(Y\mid Z)
    \big]
    + \beta\, I(X;Z).
\end{equation}
Here, the predictive term encourages $Z$ to remain sufficient for the target $Y$, while the mutual-information penalty discourages the representation from carrying unnecessary information about the raw input.

A standard variational instantiation is the variational information bottleneck:
\begin{equation}
    \mathcal{L}_{\mathrm{VIB}}
    =
    \mathbb{E}_{Z\sim q_\theta(Z\mid X)}
    \big[
        -\log p_\phi(Y\mid Z)
    \big]
    +
    \beta D_{\mathrm{KL}}\big(q_\theta(Z\mid X)\,\|\,p(Z)\big),
\end{equation}
where $p(Z)$ is a reference distribution. From the perspective of the unified framework, both the ideal and variational forms can be understood as target-driven learning augmented with a compression-oriented constraint on the representation~\citep{tishby2000ib,alemi2017vib,achille2018information}.

\subsubsection{Distribution alignment constraints}

Distribution alignment constraints encourage the learned representation to match a prescribed latent distribution. In the unified framework, these methods retain the same target-driven structure but instantiate $\Omega_{\mathrm{gen}}(Z,X)$ as a distribution-alignment term.

The most prominent example is the variational autoencoder. In this case,
\begin{equation}
    \tilde{X}=X,\qquad T(X)=X,\qquad Z\sim q_\theta(Z\mid X).
\end{equation}
A canonical task loss is the negative conditional log-likelihood of the input:
\begin{equation}
    \mathcal{L}_{\mathrm{task}}
    =
    -\log p_\phi(X\mid Z).
\end{equation}
The standard VAE objective is therefore
\begin{equation}
    \mathcal{L}_{\mathrm{VAE}}
    =
    \mathbb{E}_{Z\sim q_\theta(Z\mid X)}
    \big[
        -\log p_\phi(X\mid Z)
    \big]
    +
    \beta D_{\mathrm{KL}}\big(q_\theta(Z\mid X)\,\|\,p(Z)\big),
\end{equation}
where $p(Z)$ is typically chosen as a simple prior such as a standard Gaussian. The first term encourages the representation to preserve information sufficient for reconstruction, while the KL term regularizes the encoder-induced latent distribution toward a structured prior.

Adversarial autoencoders follow the same overall logic, but replace the KL term with an adversarial distribution-matching objective. A standard formulation is
\begin{equation}
    \min_{\theta,\phi}\max_{D_{\mathrm{adv}}}\;
    \mathbb{E}_{Z\sim q_\theta(Z\mid X)}
    \big[
        -\log p_\phi(X\mid Z)
    \big]
    +
    \lambda
    \Big(
        \mathbb{E}_{Z\sim p(Z)}[\log D_{\mathrm{adv}}(Z)]
        +
        \mathbb{E}_{Z\sim q_\theta(Z\mid X)}[\log(1-D_{\mathrm{adv}}(Z))]
    \Big),
\end{equation}
where $D_{\mathrm{adv}}$ denotes the adversarial discriminator. This keeps the reconstruction-driven task objective fixed while changing the form of the latent distribution-alignment constraint~\citep{kingma2014vae,rezende2014stochastic,makhzani2016aae}.

\subsubsection{Sparsity constraints}

Sparsity constraints favor representations in which only a small number of latent coordinates are active for a given sample. In the unified framework, this means that the target-driven objective is supplemented with a sparsity-inducing penalty on the latent code.

A canonical reconstruction-based instantiation is obtained by choosing
\begin{equation}
    \Omega_{\mathrm{gen}}(Z,X)=\|Z\|_1.
\end{equation}
Together with the standard reconstruction loss
\begin{equation}
    \mathcal{L}_{\mathrm{task}}
    =
    \|X-g_\phi(Z)\|_2^2,
\end{equation}
this yields the objective
\begin{equation}
    \min_{\theta,\phi}\;
    \mathbb{E}_{X}
    \Big[
        \|X-g_\phi(f_\theta(X))\|_2^2
        + \lambda \|f_\theta(X)\|_1
    \Big].
\end{equation}
A common alternative in sparse autoencoders is to penalize average unit activations toward a target sparsity level $\rho$, for example through
\begin{equation}
    \Omega_{\mathrm{gen}}(Z,X)
    =
    D_{\mathrm{KL}}\big(\rho\,\|\,\hat{\rho}\big),
\end{equation}
where $\hat{\rho}$ denotes the empirical activation rate. In either case, the task term preserves target-relevant information, while the regularizer favors solutions in which only a small subset of latent dimensions is active for any given observation~\citep{olshausen1996emergence,olshausen1997sparse,ng2011sparseae,cunningham2023sparse,templeton2024scaling}.

\subsubsection{Energy-based constraints}

Energy-based constraints assign lower cost to latent configurations that lie in high-preference or low-energy regions. In the unified framework, such methods choose $\Omega_{\mathrm{gen}}(Z,X)$ to reflect an energy function over the latent space.

Suppose the latent prior is defined by an energy-based model
\begin{equation}
    p_\psi(Z)\propto \exp\big(-E_\psi(Z)\big).
\end{equation}
Then the negative log-prior takes the form
\begin{equation}
    -\log p_\psi(Z)
    =
    E_\psi(Z)+\log \mathcal{Z}_\psi,
\end{equation}
where $\mathcal{Z}_\psi$ is the partition function. Up to the additive constant $\log \mathcal{Z}_\psi$, this corresponds to the energy penalty
\begin{equation}
    \Omega_{\mathrm{gen}}(Z,X)=E_\psi(Z).
\end{equation}
If the target remains reconstruction, a concrete objective is
\begin{equation}
    \mathcal{L}_{\mathrm{EB}}
    =
    \|X-g_\phi(f_\theta(X))\|_2^2
    +
    \lambda E_\psi(f_\theta(X)).
\end{equation}
Under this interpretation, the representation is constrained not only to support the task objective, but also to lie in a low-energy region of the latent space. Conceptually, this highlights that representation learning can involve preferences over which latent states are considered plausible, stable, or well-formed, in addition to what target information is preserved~\citep{lecun2006ebm,hinton2006dbn,hinton2012rbm,bengio2007greedy}.

\subsubsection{Discrete / quantized constraints}

Discrete or quantized constraints restrict the representation to pass through a finite codebook. Rather than encouraging the representation to be sparse or aligned to a continuous prior, these methods impose a finite representational format on the latent code.

A representative example is the vector-quantized variational autoencoder. In this setting,
\begin{equation}
    \tilde{X}=X,\qquad T(X)=X.
\end{equation}
The encoder first produces a continuous latent output
\begin{equation}
    z_e=f_\theta(X),
\end{equation}
which is then quantized to the nearest code in a finite codebook $\{e_k\}_{k=1}^{K}$:
\begin{equation}
    Z=e_{k^\ast},
    \qquad
    k^\ast=\arg\min_{k}\|z_e-e_k\|_2^2.
\end{equation}
A canonical reconstruction loss is
\begin{equation}
    \mathcal{L}_{\mathrm{task}}
    =
    \|X-g_\phi(Z)\|_2^2.
\end{equation}
The standard VQ objective can then be written as
\begin{equation}
    \mathcal{L}_{\mathrm{VQ}}
    =
    \|X-g_\phi(Z)\|_2^2
    +
    \|\mathrm{sg}[z_e]-e_{k^\ast}\|_2^2
    +
    \beta\|z_e-\mathrm{sg}[e_{k^\ast}]\|_2^2,
\end{equation}
where $\mathrm{sg}[\cdot]$ denotes the stop-gradient operator. From the standpoint of the unified framework, the task term encourages the representation to preserve information sufficient for reconstruction, while the additional terms force the representation through a discrete, finite-capacity bottleneck. This makes VQ-VAE a clear example of representation learning shaped not only by what target signal should be preserved, but also by what representational format is admissible~\citep{oord2017vqvae,razavi2019vqvae2,esser2021vqgan,yu2023language,esser2021taming}.

\subsection{Causal representation learning: causal constraints}
\label{app:crl_details}

The previous subsection considered generic additional constraints that shape the latent space without requiring the learned coordinates to correspond to causal variables. Causal representation learning imposes a stronger requirement: the learned representation should align with the latent variables and mechanisms of an underlying structural causal model, so that it supports intervention-aware reasoning rather than only prediction or reconstruction~\citep{pearl2009causality,peters2017elements,scholkopf2021toward}.

Let $Z^*=(Z_1^*,\dots,Z_D^*)$ denote the true latent causal variables generated by a structural causal model
\begin{equation}
    Z_j^* := m_j\big(Z_{\operatorname{pa}_G(j)}^*, \epsilon_j\big),
    \qquad j=1,\dots,D,
    \qquad \epsilon_i \perp \epsilon_j \text{ for } i\neq j,
\end{equation}
where $G$ is a directed acyclic graph and $\operatorname{pa}_G(j)$ denotes the parents of node $j$ in $G$. The observation is produced through an unknown measurement or mixing mechanism
\begin{equation}
    X = g^*(Z^*).
\end{equation}
CRL seeks to learn a representation $Z=f_\theta(\tilde{X})$ that corresponds to these latent causal variables, or at least to causally meaningful latent modules. The key requirement is not merely that $Z$ be predictive, compressive, or statistically factorized, but that it admit an intervention-consistent interpretation: changing one learned factor should correspond to changing one underlying causal factor, with effects propagating according to the latent causal structure. In this sense, CRL is stronger than ordinary disentanglement, since causal variables need not be marginally independent and a factorized representation is not necessarily causal~\citep{scholkopf2021toward,yang2021causalvae}.

\subsubsection{Causal representations and identifiability}

Because both the true latent causal variables $Z^*$ and the observation map $g^*$ are hidden, exact recovery is generally not possible without assumptions. Instead, causal recovery is typically defined up to an equivalence class.

In the simplest and most interpretable case, the learned representation is identifiable up to permutation and component-wise invertible transformations, meaning that there exist a permutation $\pi$ of $\{1,\dots,D\}$ and invertible scalar functions $\psi_1,\dots,\psi_D$ such that
\begin{equation}
    Z_j = \psi_j\!\big(Z_{\pi(j)}^*\big),
    \qquad j=1,\dots,D.
\end{equation}
This means that each learned coordinate corresponds to one true latent causal variable, up to relabeling and reparameterization.

A weaker notion allows recovery only at the level of blocks. Let $\mathcal{B}=\{B_1,\dots,B_K\}$ be a partition of the latent coordinates. Then one may only have
\begin{equation}
    Z_{B_k} = \Psi_k\!\big(Z_{B_{\pi(k)}}^*\big),
    \qquad k=1,\dots,K,
\end{equation}
for invertible blockwise maps $\Psi_k$. In this case, the learned representation identifies causal modules rather than individual scalar variables.

The essential point is therefore not exact equality between $Z$ and $Z^*$, but that the remaining ambiguity should be limited enough for learned coordinates or blocks to retain a causal interpretation and remain compatible with intervention semantics~\citep{hyvarinen2019nonlinear,khemakhem2020ivae,ahuja2023interventional}.

\subsubsection{CRL as representation learning with causal constraints}

Under the unified formulation, causal representation learning can be written as
\begin{equation}
\label{eq:app_crl_objective}
    Z=f_\theta(\tilde{X}),\qquad
    \min_{\theta,\phi}\;
    \mathbb{E}_{X,\tilde{X}\sim v(X)}
    \Big[
        \mathcal{L}_{\mathrm{task}}\big(g_\phi(Z),\,T(X)\big)
        + \lambda\,\Omega_{\mathrm{causal}}(Z,X)
    \Big].
\end{equation}
The task component $\mathcal{L}_{\mathrm{task}}\big(g_\phi(Z),T(X)\big)$ plays the same formal role as in standard representation learning. Many generative CRL methods use reconstruction, while other methods may use prediction, contrastive learning, masked prediction, or temporal objectives. The defining ingredient is the causal constraint $\Omega_{\mathrm{causal}}(Z,X)$, which biases the learned representation toward one that is compatible with latent causal structure.

Unlike generic constraints that encourage compression, sparsity, prior matching, energy-based preferences, or quantization, the causal constraint is intended to make the latent representation consistent with causal structure. Depending on the setting, it may encourage properties such as modularity of latent mechanisms, invariance of non-intervened factors across environments, localization of intervention effects, consistency across paired observations, or compatibility with a latent causal graph. The precise mathematical form of $\Omega_{\mathrm{causal}}(Z,X)$ depends on what assumptions and supervision are available.

From the perspective of the unified formulation, causal representation learning is a special case of representation learning with additional constraints. The observed view $\tilde{X}$ and target signal $T(X)$ may remain the same as in supervised, reconstruction-based, or multi-view learning, but the admissible representations are further restricted by the requirement that they be causally meaningful.

\subsubsection{Sparse mechanism constraints}

Sparse mechanism constraints assume that the latent causal mechanisms are sparse: each learned factor should depend only on a small number of parents or auxiliary variables. Let $\hat{E}$ denote an estimated edge-weight structure over latent variables. A natural causal constraint is
\begin{equation}
    \Omega_{\mathrm{causal}}(Z,X)=\|\hat{E}\|_1.
\end{equation}
This biases the learned representation toward mechanisms that are local and modular rather than densely entangled. Mechanism sparsity regularization is a representative example of this principle: it jointly learns latent variables and a sparse causal graphical structure, often through regularization on latent dependencies~\citep{lachapelle2022mechanism,zhang2024multiple,zheng2022sparsity,idol,el2024toward}.

\subsubsection{Conditional-prior constraints}

Conditional-prior constraints constrain the distribution of the latent code itself. In the static setting, one introduces a structured prior conditioned on auxiliary information:
\begin{equation}
    p_\eta(Z\mid U),
\end{equation}
where $U$ may denote a domain index, label, temporal indicator, or other auxiliary variable. A common constraint regularizes the encoder by
\begin{equation}
    \Omega_{\mathrm{causal}}(Z,X)
    =
    D_{\mathrm{KL}}\!\big(q_\theta(Z\mid X)\,\|\,p_\eta(Z\mid U)\big).
\end{equation}
The identifying signal comes from sufficient variability of $U$, which changes the latent distribution in a structured way. In temporal settings, the same idea becomes a structured transition prior:
\begin{equation}
    \Omega_{\mathrm{causal}}(Z,X)
    =
    D_{\mathrm{KL}}\!\big(
        q_\theta(Z_t\mid X_{\le t})
        \,\|\,
        p_\eta(Z_t\mid Z_{<t},U_t)
    \big).
\end{equation}
One may further decompose the prior into stable and changing components across environments. Thus, the causal constraint is expressed by aligning the inferred latent code with a structured conditional or transition prior rather than with an unconditional isotropic prior. This viewpoint underlies iVAE-style identification, domain-adaptation formulations with invariant and changing latent subspaces, and temporal CRL methods such as TDRL~\citep{khemakhem2020ivae,kong2022partialdisentanglement,yaotdrl,pcl,shen2022weakly}.

\subsubsection{Invariance and intervention constraints}

Invariance and intervention constraints can be understood through an invariance principle: for each related pair of observations or environments $(v_1,v_2)$, there exists a subset $A\subseteq[D]$ of latent coordinates whose relevant statistic is preserved across the pair. This yields the generic causal regularizer
\begin{equation}
    \Omega_{\mathrm{causal}}(Z,X)
    =
    d\!\big(
        \mathcal{S}^{(v_1)}_{A}(Z^{(v_1)}),
        \mathcal{S}^{(v_2)}_{A}(Z^{(v_2)})
    \big),
\end{equation}
where $d(\cdot,\cdot)$ measures discrepancy and $\mathcal{S}^{(v_1)}_{A}$ denotes a problem-dependent invariant statistic on the subset $A$ for the view, observation, or environment $v_1$. Depending on the setting, this statistic may encode sample-level equality, marginal invariance, score invariance, transition invariance, or other preserved structure across related data pockets.

Intervention-based CRL can be viewed as a special case of the same principle. When $(v_1,v_2)$ are related by an intervention with target $I$, and when a latent causal graph $G$ is available, one common choice is
\begin{equation}
    A=[D]\setminus\operatorname{De}_G(I),
\end{equation}
where $\operatorname{De}_G(I)$ denotes the descendants of $I$ in $G$. The objective then matches the corresponding invariant statistic on that subset. Some methods additionally combine this invariance term with sparsity or localization penalties so that pair-specific variation is concentrated on the complementary coordinates $[D]\setminus A$. Thus, different CRL categories differ mainly in how the invariant subset $A$ is specified and how $\mathcal{S}^{(v_1)}_{A}$ is instantiated, rather than in the underlying learning principle itself~\citep{locatello2020weakly,yao2024unifying,ahuja2023interventional,lippe2022citris,varici2024general,varici2025score,welch2024identifiability}.

\subsubsection{Functional constraints}

Functional constraints impose causal inductive biases through the generative process or the structure of the latent space. These biases can take many forms, including object-centric generators, modular decoders, and additive decoders~\citep{locatello2020slotattention,ren2026causal,li2025towards}. As one representative example, additive decoders encourage a structured decomposition of the generator by partitioning the representation into blocks. Let $\mathcal{B}=\{B_1,\dots,B_K\}$ be a partition of the latent coordinates and write
\begin{equation}
    Z=(Z_{B_1},\dots,Z_{B_K}).
\end{equation}
Instead of learning an unrestricted decoder, these methods constrain the decoder to combine block-specific contributions:
\begin{equation}
    g_\phi(Z)
    =
    \sum_{k=1}^{K}
    \alpha_\phi^{(k)}(Z)\,g_\phi^{(k)}(Z_{B_k}),
\end{equation}
where each branch $g_\phi^{(k)}$ depends only on one latent block and $\alpha_\phi^{(k)}(Z)$ controls the contribution of that branch, often with $\sum_{k=1}^{K}\alpha_\phi^{(k)}(Z)=1$. The causal bias therefore comes from forcing the observation to be generated from separate block-specific contributions rather than from an unrestricted entangled decoder.

Object-centric decoders and additive decoder models can both be viewed through this functional-constraint perspective. Object-centric models use separate slots or blocks to generate parts of the observation, while additive decoder models make the blockwise contributions explicit and support identifiability and Cartesian-product extrapolation. Although the precise mixing mechanism differs across implementations, these approaches share the same high-level decoder-structure bias~\citep{locatello2020slotattention,lachapelle2023additive,ren2026causal,li2025towards}.

%% file: appendix/3-experiment_details.tex
\section{Experiment Details}
\label{section:experiment_details}

\subsection{Task-Conditioned TDRL Implementation}
\label{subsec:tdrl_task_implementation}

\paragraph{Model architecture.}
We implement all video task variants on top of the same temporally regularized
latent representation model. The purpose of this design is to keep the
representation backbone fixed and only change the task objective. Therefore,
differences across methods can be interpreted as the effect of different task
signals rather than changes in model capacity.

Given a video clip
\begin{equation}
    X=(x_1,\dots,x_T),
\end{equation}
where each $x_t$ denotes the latent observation at time $t$, a frame-wise
encoder first maps each frame independently to a feature representation:
\begin{equation}
    h_t = E_{\psi}(x_t).
\end{equation}
The sequence of frame features $h_{1:T}$ is then mapped to a sequence of
stochastic causal latents $z_{1:T}$. For each time step, the inference network
parameterizes a diagonal Gaussian posterior:
\begin{equation}
    q_{\theta}(z_t\mid h_t)
    =
    \mathcal{N}
    \Big(
        z_t;
        \mu_{\theta}(h_t),
        \mathrm{diag}(\exp(\ell_{\theta}(h_t)))
    \Big),
\end{equation}
where $\mu_{\theta}$ and $\ell_{\theta}$ are the mean and log-variance heads.
Both heads have the same architecture and are shared across all task variants.
The latent variable is sampled using the reparameterization trick:
\begin{equation}
    z_t
    =
    \mu_{\theta}(h_t)
    +
    \epsilon_t
    \odot
    \exp\!\left(\frac{1}{2}\ell_{\theta}(h_t)\right),
    \qquad
    \epsilon_t\sim\mathcal{N}(0,I).
\end{equation}
During evaluation, we use the deterministic posterior mean by disabling random
sampling.

The decoder is history-conditioned. For a fixed lag order $L$, we construct a
temporal window of latent states
\begin{equation}
    s_t=[z_{t-L},z_{t-L+1},\dots,z_t].
\end{equation}
For the first few frames, missing history entries are padded with zeros. The
history vector $s_t$ is passed to a decoder head to produce a predicted feature:
\begin{equation}
    \hat h_t = D_{\phi}(s_t).
\end{equation}
A frame decoder then maps $\hat h_t$ back to the observation latent space:
\begin{equation}
    \hat x_t = G_{\psi}(\hat h_t).
\end{equation}
Thus, the common computational path for all video tasks is
\begin{equation}
    X
    \longrightarrow
    h_{1:T}
    \longrightarrow
    z_{1:T}
    \longrightarrow
    \hat h_{1:T}
    \longrightarrow
    \hat X.
\end{equation}
The frame encoder, causal posterior heads, history-window construction, decoder
head, and frame decoder are kept unchanged for all task variants. In the code,
each method is implemented as a separate model file, but the shared structural
components are kept identical for controlled comparison.

\paragraph{Fixed representation constraint and training objective.}
The temporal constraint follows the TDRL-style transition prior. The first few
latent states are regularized by a standard Gaussian prior:
\begin{equation}
    p(z_t)=\mathcal{N}(0,I),
    \qquad t\le L.
\end{equation}
This gives the initial KL term
\begin{equation}
    \mathcal{L}_{\mathrm{init}}
    =
    \sum_{t=1}^{L}
    \mathrm{KL}
    \Big(
        q_{\theta}(z_t\mid h_t)
        \,\|\, \mathcal{N}(0,I)
    \Big).
\end{equation}
For later time steps, the prior is defined through a nonlinear temporal
transition model. The transition module transforms the latent sequence into
residual variables and returns a log-Jacobian determinant:
\begin{equation}
    (r_{L+1:T}, \log|\det J|)
    =
    F_{\eta}(z_{1:T}).
\end{equation}
The residuals are modeled using a standard Gaussian base density. By the
change-of-variables formula, the temporal prior likelihood can be written as
\begin{equation}
    \log p_{\eta}(z_{L+1:T}\mid z_{1:L})
    =
    \sum_{t=L+1}^{T}
    \log \mathcal{N}(r_t;0,I)
    +
    \log|\det J|.
\end{equation}
The future temporal KL is estimated by comparing the posterior log-probability
with this transition-prior log-probability:
\begin{equation}
    \mathcal{L}_{\mathrm{future}}
    =
    \frac{1}{T-L}
    \mathbb{E}_{q_{\theta}}
    \left[
        \log q_{\theta}(z_{L+1:T}\mid X)
        -
        \log p_{\eta}(z_{L+1:T}\mid z_{1:L})
    \right].
\end{equation}
The temporal regularization term is
\begin{equation}
    \Omega_{\mathrm{TDRL}}
    =
    \beta \mathcal{L}_{\mathrm{init}}
    +
    \gamma \mathcal{L}_{\mathrm{future}}.
\end{equation}
This constraint is shared by all task variants. It encourages the latent
representation to follow a temporally structured transition model rather than
behaving as independent frame-wise features.

In addition to the TDRL constraint, we keep two auxiliary losses fixed across
all methods. The first one reconstructs the frame feature from the decoded
feature:
\begin{equation}
    \mathcal{L}_{\mathrm{latent}}
    =
    \frac{1}{T}
    \sum_{t=1}^{T}
    \|\hat h_t-h_t\|_2^2.
\end{equation}
This term stabilizes the decoder head and aligns the decoded feature with the
frame-wise representation space. The second one matches local temporal
differences in feature space:
\begin{equation}
    \mathcal{L}_{\mathrm{delta}}
    =
    \frac{1}{T-1}
    \sum_{t=1}^{T-1}
    \left\|
        (\hat h_{t+1}-\hat h_t)
        -
        (h_{t+1}-h_t)
    \right\|_2^2.
\end{equation}
This term encourages the decoded features to preserve short-term temporal
variation. Importantly, these two losses are fixed regularization terms and are
not used to define different task variants.

The complete objective for a task-specific loss
$\mathcal{L}_{\mathrm{task}}$ is
\begin{equation}
    \mathcal{L}_{\mathrm{total}}
    =
    \lambda_{\mathrm{task}}\mathcal{L}_{\mathrm{task}}
    +
    \lambda_{\mathrm{latent}}\mathcal{L}_{\mathrm{latent}}
    +
    \lambda_{\mathrm{delta}}\mathcal{L}_{\mathrm{delta}}
    +
    \Omega_{\mathrm{TDRL}}.
\end{equation}
Therefore, each task variant only changes $\mathcal{L}_{\mathrm{task}}$, while
the representation model, temporal constraint, and auxiliary regularizers remain
fixed. For code compatibility, the task-specific loss is stored under the key
\texttt{recon\_loss}, even when the task is not ordinary reconstruction.

\paragraph{Reconstruction (TDRL).}
Reconstruction (TDRL) is the standard video baseline. The decoded observation
$\hat x_t$ is supervised by the current input $x_t$:
\begin{equation}
    \mathcal{L}_{\mathrm{task}}^{\mathrm{recon}}
    =
    \frac{1}{T}
    \sum_{t=1}^{T}
    \|\hat x_t-x_t\|_2^2.
\end{equation}
This objective preserves information needed to reconstruct the current frame and
serves as the reference task in the video setting.

\paragraph{Contrastive Learning.}
Contrastive Learning uses a CPC-style objective. The decoded feature is used as
a query,
\begin{equation}
    q_t =
    \frac{\hat h_t}{\|\hat h_t\|_2},
\end{equation}
and the next-frame feature is used as the positive key,
\begin{equation}
    k_{t+1}
    =
    \frac{\mathrm{sg}(h_{t+1})}
    {\|\mathrm{sg}(h_{t+1})\|_2},
\end{equation}
where $\mathrm{sg}(\cdot)$ denotes stop-gradient. Flattening all valid time
indices and batch elements into an index set $\mathcal{I}$, the task loss is
\begin{equation}
    \mathcal{L}_{\mathrm{task}}^{\mathrm{CPC}}
    =
    -
    \frac{1}{|\mathcal{I}|}
    \sum_{i\in\mathcal{I}}
    \log
    \frac{
        \exp(q_i^\top k_i/\tau)
    }{
        \sum_{j\in\mathcal{I}}
        \exp(q_i^\top k_j/\tau)
    }.
\end{equation}
This objective encourages the current latent history to identify the correct
future feature among in-batch candidates.

\paragraph{Next-Frame Prediction.}
Next-Frame Prediction trains the model to predict future observations from the
current latent history. The task loss is
\begin{equation}
    \mathcal{L}_{\mathrm{task}}^{\mathrm{next}}
    =
    \frac{1}{T-1}
    \sum_{t=1}^{T-1}
    \|\hat x_t-x_{t+1}\|_2^2.
\end{equation}
This objective changes the target from the current frame to the next frame,
encouraging $z_t$ to capture information useful for short-term temporal
prediction.

\paragraph{Mid-Latent Reconstruction.}
Mid-Latent Reconstruction moves the reconstruction target from the observation
space to the intermediate feature space. Instead of supervising $\hat x_t$, the
task directly matches the decoded feature $\hat h_t$ to the encoder feature
$h_t$:
\begin{equation}
    \mathcal{L}_{\mathrm{task}}^{\mathrm{mid}}
    =
    \frac{1}{T}
    \sum_{t=1}^{T}
    \|\hat h_t-\mathrm{sg}(h_t)\|_2^2.
\end{equation}
This objective tests whether reconstructing an intermediate representation,
rather than the frame itself, provides a better task signal for learning causal
latents.

\paragraph{Prototype-based Learning.}
Prototype-based Learning forms online cluster assignments from the learned
latent representation. A representation $r_i$ is obtained from either a
frame-level latent or a sequence-level aggregation of $z_{1:T}$. Batch
prototypes $c_k$ are estimated online, and each representation is assigned to
its nearest prototype:
\begin{equation}
    a_i
    =
    \arg\min_k
    \|\mathrm{sg}(r_i)-c_k\|_2^2.
\end{equation}
The task loss is a prototype classification loss:
\begin{equation}
    \mathcal{L}_{\mathrm{task}}^{\mathrm{proto}}
    =
    -
    \frac{1}{N}
    \sum_{i=1}^{N}
    \log
    \frac{
        \exp(\bar r_i^\top \bar c_{a_i}/\tau_p)
    }{
        \sum_k
        \exp(\bar r_i^\top \bar c_k/\tau_p)
    },
\end{equation}
where $\bar r_i$ and $\bar c_k$ denote normalized representations and
prototypes. This objective encourages the latent space to form stable state or
regime clusters.

\paragraph{Masked Reconstruction.}
Masked Reconstruction provides the encoder with a masked view of the input:
\begin{equation}
    \tilde x_t=m_t\odot x_t,
\end{equation}
where $m_t$ is a binary mask. The model is trained to reconstruct the masked-out
part of the original observation:
\begin{equation}
    \mathcal{L}_{\mathrm{task}}^{\mathrm{mask}}
    =
    \frac{
        \sum_{t=1}^{T}
        \|(1-m_t)\odot(\hat x_t-x_t)\|_2^2
    }{
        \sum_{t=1}^{T}
        \|1-m_t\|_1
        +
        \epsilon
    }.
\end{equation}
This objective encourages the representation to recover missing information
from the visible temporal context. During evaluation, we use the clean input
view so that the measured latent representation is not affected by stochastic
masking noise.

Across all video experiments, the representation used for evaluation is the
learned causal latent sequence $z_{1:T}$. Since the model architecture,
temporal constraint, and auxiliary losses are fixed, the comparison isolates the
effect of the chosen task signal on the quality of the learned temporal causal
representation.

\subsection{Task-Conditioned iMSDA Implementation}
\label{subsec:imsda_task_implementation}

\paragraph{Model architecture.}
All image task variants use the same iMSDA backbone and differ only in the task
objective. Given an input image $X$ and its domain or view index $u$, a frozen
ResNet18 feature extractor first maps the image to a feature vector
\begin{equation}
    x = \varphi(X).
\end{equation}
The feature $x$ is then passed through an encoder network to parameterize a
diagonal Gaussian posterior:
\begin{equation}
    q_\theta(z\mid x)
    =
    \mathcal{N}
    \big(
        z;
        \mu_\theta(x),
        \mathrm{diag}(\sigma_\theta^2(x))
    \big).
\end{equation}
During training, the latent code $z$ is sampled using the reparameterization
trick. The latent variable is partitioned into a content part and a
style-specific part:
\begin{equation}
    z = (z_c, z_s),
    \qquad
    z_c \in \mathbb{R}^{c}, \quad z_s \in \mathbb{R}^{s}.
\end{equation}
The content part is kept unchanged, while the style part is transformed by a
domain-conditioned invertible flow:
\begin{equation}
    \tilde z_s = f_u(z_s),
    \qquad
    \tilde z = (z_c,\tilde z_s).
\end{equation}
Here the flow parameters are generated from an embedding of the view index $u$.
The content representation used for evaluation is the content part
$\tilde z_c$. A decoder $g_\omega$ maps the sampled latent $z$ back to the
ResNet feature space:
\begin{equation}
    \hat x = g_\omega(z).
\end{equation}
Therefore, reconstruction in our implementation always refers to
backbone-feature reconstruction rather than pixel-level image reconstruction.

\paragraph{Fixed representation constraint and training objective.}
Across all image tasks, the representation constraint is kept fixed. The first
constraint is a capacity-controlled variational KL term. For each domain, we
estimate
\begin{equation}
    \mathrm{KL}
    =
    \mathbb{E}_{q_\theta(z\mid x)}
    \left[
        \log q_\theta(z\mid x)
        -
        \log p(\tilde z)
        -
        \log\left|\det J_{f_u}\right|
    \right],
\end{equation}
where $p(\tilde z)$ is a standard Gaussian prior and
$\log|\det J_{f_u}|$ is the log-determinant of the domain-conditioned flow.
Following a capacity schedule, the KL penalty is written as
\begin{equation}
    \mathcal{L}_{\mathrm{KL}}
    =
    \beta
    \left|
        \mathrm{KL} - C(t)
    \right|,
    \qquad
    C(t)=
    \min\left(C_{\max}, \frac{t}{T_{\mathrm{stop}}}C_{\max}\right).
\end{equation}
The second constraint regularizes the transformed style code $\tilde z_s$
toward a standard Gaussian:
\begin{equation}
    \mathcal{L}_{\mathrm{style}}
    =
    -\log p(\tilde z_s).
\end{equation}
The total loss is
\begin{equation}
    \mathcal{L}_{\mathrm{total}}
    =
    \lambda_{\mathrm{vae}}
    \left(
        \mathcal{L}_{\mathrm{task}}
        +
        \mathcal{L}_{\mathrm{KL}}
    \right)
    +
    \lambda_{\mathrm{gauss}}
    \mathcal{L}_{\mathrm{style}}.
\end{equation}
This objective is fixed in all image experiments. We only replace
$\mathcal{L}_{\mathrm{task}}$ to instantiate different representation-learning
tasks under the same iMSDA constraint.

\paragraph{Reconstruction(iMSDA).}
Reconstruction(iMSDA) is the standard image baseline. The model observes the
original image $X$, extracts its backbone feature $\varphi(X)$, encodes it into
$z$, and decodes $z$ back to the same feature space:
\begin{equation}
    \tilde X = X,
    \qquad
    T(X)=\varphi(X).
\end{equation}
The task loss is
\begin{equation}
    \mathcal{L}_{\mathrm{task}}^{\mathrm{iMSDA}}
    =
    \|g_\omega(z(X))-\varphi(X)\|_2^2.
\end{equation}
This task keeps the original iMSDA feature-reconstruction objective unchanged.

\paragraph{Denoising Reconstruction.}
Denoising Reconstruction changes only the observed view. The encoder receives a
corrupted image $c(X)$, while the target remains the clean backbone feature:
\begin{equation}
    \tilde X = c(X),
    \qquad
    T(X)=\varphi(X).
\end{equation}
The task loss is
\begin{equation}
    \mathcal{L}_{\mathrm{task}}^{\mathrm{denoise}}
    =
    \|g_\omega(z(c(X)))-\varphi(X)\|_2^2.
\end{equation}
This objective encourages the representation to retain information that is
stable under input corruption.

\paragraph{Masked Reconstruction.}
Masked Reconstruction gives the encoder only a partially observed image. Let
$M$ denote the masked region and $\bar M$ the visible region:
\begin{equation}
    \tilde X = X_{\bar M},
    \qquad
    T(X)=\varphi(X).
\end{equation}
The task loss is
\begin{equation}
    \mathcal{L}_{\mathrm{task}}^{\mathrm{mask}}
    =
    \|g_\omega(z(X_{\bar M}))-\varphi(X)\|_2^2.
\end{equation}
This task encourages the representation to recover the clean feature from
partial observations.

\paragraph{Cross-view Prediction.}
Cross-view Prediction uses two different views $v_i(X)$ and $v_j(X)$ of the
same sample. The model encodes one view and predicts the backbone feature of
the other view:
\begin{equation}
    \tilde X = v_i(X),
    \qquad
    T(X)=\varphi(v_j(X)),
    \qquad i\neq j.
\end{equation}
We use a symmetric feature prediction loss:
\begin{equation}
    \mathcal{L}_{\mathrm{task}}^{\mathrm{cross}}
    =
    \frac{1}{2}
    \left(
        \|g_\omega(z(v_i(X)))-\varphi(v_j(X))\|_2^2
        +
        \|g_\omega(z(v_j(X)))-\varphi(v_i(X))\|_2^2
    \right).
\end{equation}
This objective encourages the learned representation to capture content shared
across views.

\paragraph{Contrastive Learning.}
Contrastive Learning applies a multi-view InfoNCE objective to the content
branch. For two views $v_i(X)$ and $v_j(X)$ of the same sample, we extract
their content codes:
\begin{equation}
    \tilde z_c^{(i)}=\tilde z_c(v_i(X)),
    \qquad
    \tilde z_c^{(j)}=\tilde z_c(v_j(X)).
\end{equation}
A projection head maps them to embeddings $H_i$ and $H_j$. The task loss is
\begin{equation}
    \mathcal{L}_{\mathrm{task}}^{\mathrm{contrast}}
    =
    \frac{1}{2}
    \left[
    \mathrm{CE}
    \left(
        \frac{H_iH_j^\top}{\tau},
        y
    \right)
    +
    \mathrm{CE}
    \left(
        \frac{H_jH_i^\top}{\tau},
        y
    \right)
    \right],
\end{equation}
where $y$ indexes the matching same-sample pairs in the batch and $\tau$ is the
temperature parameter. This objective pulls together content representations of
the same sample across views and pushes apart representations from different
samples.

\paragraph{Prototype-based Learning.}
Prototype-based Learning applies a clustering-style objective to the content
branch. Given two views of the same sample, we project their content codes and
assign them to learned prototypes. Let $q_i$ and $q_j$ denote balanced prototype
assignments from the two views. The task loss is
\begin{equation}
    \mathcal{L}_{\mathrm{task}}^{\mathrm{proto}}
    =
    -\frac{1}{2}
    \left[
        \sum_k q_j^{(k)}
        \log p_\chi(k\mid \tilde z_c(v_i(X)))
        +
        \sum_k q_i^{(k)}
        \log p_\chi(k\mid \tilde z_c(v_j(X)))
    \right].
\end{equation}
The assignments are balanced using Sinkhorn normalization. This task encourages
the content representation to form stable view-invariant prototypes.

Across all image experiments, the representation used for evaluation is the
learned content representation $\tilde z_c$. Since the iMSDA backbone,
domain-conditioned constraint, and training protocol are fixed, the comparison
isolates the effect of the chosen task signal on the quality of the learned
image representation.

\subsection{Task Variants under a Decoder-Sparsity Constraint}
\label{subsec:sparsity_task_variants}

\subsubsection{Task ablation under decoder sparsity}

We further evaluate whether the task objective also matters when the main
representation constraint is decoder sparsity. We use a VAE with a decoder
Jacobian sparsity penalty as the baseline, denoted as
\texttt{Reconstruction (VAE+Sparsity)}. Starting from this baseline, we keep the
VAE KL regularization and the decoder-sparsity constraint fixed, and only
replace the task objective. This ablation tests whether different task signals
can improve representation quality under the same sparsity-based constraint.

\begin{table}[t]
\centering
\caption{Comparison of task choices under the VAE+sparsity setting. All variants use the same KL and decoder-sparsity constraints, while only the task objective is changed.}
\label{tab:sparsity_task_comparison}
\begin{tabular}{l c c}
\hline
\textbf{Task} & \textbf{MCC} & $\boldsymbol{R^2}$ \\
\hline
Reconstruction (VAE+Sparsity) & 0.40 & 0.92 \\
Denoising Reconstruction      & 0.47 & 0.92 \\
Masked Reconstruction         & 0.43 & 0.88 \\
Multi-view Prediction         & 0.46 & 0.84 \\
\hline
\end{tabular}
\end{table}

Table~\ref{tab:sparsity_task_comparison} shows that the task objective remains
important even when the sparsity constraint is fixed. Denoising Reconstruction
achieves the best MCC, improving the baseline from $0.40$ to $0.47$ while
maintaining the same $R^2$. Multi-view Prediction also improves MCC to $0.46$,
suggesting that predicting a shared multi-view target encourages more
view-invariant representations. Masked Reconstruction gives a smaller
improvement in MCC but reduces $R^2$. Overall, these results support the same
conclusion as the main experiments: the constraint and the task objective play
complementary roles, and a better-matched task can improve representation
quality under a fixed constraint.

\subsubsection{Implementation details}

All sparsity-constrained variants use the same VAE backbone. Given an image
$X$, a frozen visual backbone first extracts a feature vector
\begin{equation}
    x = \varphi(X).
\end{equation}
The encoder maps $x$ to a diagonal Gaussian posterior,
\begin{equation}
    q_\theta(z\mid x)
    =
    \mathcal{N}
    \big(
        z;
        \mu_\theta(x),
        \operatorname{diag}(\sigma_\theta^2(x))
    \big),
\end{equation}
and the decoder maps the sampled latent code back to the feature space:
\begin{equation}
    \hat{x}=g_\omega(z).
\end{equation}
Thus, reconstruction in this experiment refers to backbone-feature
reconstruction rather than pixel-level image reconstruction.

The total objective is
\begin{equation}
    \mathcal{L}_{\mathrm{total}}
    =
    \lambda_{\mathrm{vae}}
    \left(
        \mathcal{L}_{\mathrm{task}}
        +
        \mathcal{L}_{\mathrm{KL}}
    \right)
    +
    \lambda_{\mathrm{sp}}
    \mathcal{L}_{\mathrm{sp}},
\end{equation}
where $\mathcal{L}_{\mathrm{KL}}$ is the standard VAE KL term and
$\mathcal{L}_{\mathrm{sp}}$ is the decoder-sparsity penalty. The sparsity term
is defined on the decoder Jacobian:
\begin{equation}
    \mathcal{L}_{\mathrm{sp}}
    =
    \frac{1}{B}
    \sum_{i=1}^{B}
    \left\|
        \frac{\partial g_\omega(z_i)}{\partial z_i}
    \right\|_1.
\end{equation}
This penalty encourages each latent dimension to affect only a sparse subset of
the decoded feature dimensions. Across all variants,
$\mathcal{L}_{\mathrm{KL}}$ and $\mathcal{L}_{\mathrm{sp}}$ are kept unchanged;
only $\mathcal{L}_{\mathrm{task}}$ is modified.

\paragraph{Reconstruction (VAE+Sparsity).}
Reconstruction (VAE+Sparsity) is the baseline task. The encoder receives the
clean image feature and the decoder reconstructs the same feature:
\begin{equation}
    \mathcal{L}_{\mathrm{task}}^{\mathrm{recon}}
    =
    \frac{1}{B}
    \sum_{i=1}^{B}
    \left\|
        g_\omega(z_i) - \varphi(X_i)
    \right\|_2^2,
    \qquad
    z_i\sim q_\theta(z\mid \varphi(X_i)).
\end{equation}
This task keeps the standard VAE feature-reconstruction objective while adding
the decoder-sparsity constraint.

\paragraph{Denoising Reconstruction.}
Denoising Reconstruction changes only the observed view. The encoder receives a
corrupted image $c(X_i)$, while the target remains the clean backbone feature:
\begin{equation}
    \mathcal{L}_{\mathrm{task}}^{\mathrm{denoise}}
    =
    \frac{1}{B}
    \sum_{i=1}^{B}
    \left\|
        g_\omega(z_i^{c}) - \varphi(X_i)
    \right\|_2^2,
    \qquad
    z_i^{c}\sim q_\theta(z\mid \varphi(c(X_i))).
\end{equation}
This objective encourages the representation to preserve information that is
stable under input corruption.

\paragraph{Masked Reconstruction.}
Masked Reconstruction gives the encoder a partially observed image
$X_{i,\bar{M}}$, where $\bar{M}$ denotes the visible region. The decoder is
trained to recover the full clean feature:
\begin{equation}
    \mathcal{L}_{\mathrm{task}}^{\mathrm{mask}}
    =
    \frac{1}{B}
    \sum_{i=1}^{B}
    \left\|
        g_\omega(z_i^{M}) - \varphi(X_i)
    \right\|_2^2,
    \qquad
    z_i^{M}\sim q_\theta(z\mid \varphi(X_{i,\bar{M}})).
\end{equation}
This task encourages the latent representation to infer missing information
from partial observations.

\paragraph{Multi-view Prediction.}
Multi-view Prediction uses multiple views of the same instance. For each
instance, we construct a canonical target feature by averaging the backbone
features across its views:
\begin{equation}
    \bar{x}_i
    =
    \frac{1}{V}
    \sum_{v=1}^{V}
    \varphi(X_i^{(v)}).
\end{equation}
Each view is then trained to predict the same canonical target:
\begin{equation}
    \mathcal{L}_{\mathrm{task}}^{\mathrm{multi}}
    =
    \frac{1}{BV}
    \sum_{i=1}^{B}
    \sum_{v=1}^{V}
    \left\|
        g_\omega(z_i^{(v)}) - \bar{x}_i
    \right\|_2^2,
    \qquad
    z_i^{(v)}\sim q_\theta(z\mid \varphi(X_i^{(v)})).
\end{equation}
This objective encourages different views of the same instance to share a
common representation target. The train--test split is fixed before
constructing multi-view groups, and grouping is applied only within the
training split. Thus, all variants use the same data split and evaluation
protocol.